\documentclass[journal]{ieeetran}

\usepackage{amsmath,amssymb,bm} % basic
\usepackage{epsfig}             % for postscript graphics files
\usepackage{textcomp}           % for Greek lowercase µ, degree symbol °, etc.
\usepackage{multirow}           % for table item spanning multiple rows
\usepackage{array}              % for table spacing
\usepackage{xcolor}             % for adding comments in another color
\usepackage{tikz}               % for drawing block diagrams
\usepackage{mathtools}          % for symbol definition
\usepackage{color,soul}         % for highlighting modified text
\usepackage{tikz}               % for flowcharts
\usetikzlibrary{shapes.geometric, arrows} % for flowcharts
\tikzstyle{io} = [trapezium, trapezium left angle=70, trapezium right angle=110, minimum width=3cm, minimum height=1cm, text width = 2.5cm, text centered, draw=black]
\tikzstyle{process} = [rectangle, minimum width=3cm, minimum height=1cm, text width = 3cm, text centered, draw=black]
\tikzstyle{decision} = [diamond, aspect=2, text centered, draw=black]
\tikzstyle{arrow} = [thick,->,>=stealth]

\makeatletter
\renewcommand*\env@matrix[1][*\c@MaxMatrixCols c]{%
  \hskip -\arraycolsep
  \let\@ifnextchar\new@ifnextchar
  \array{#1}}
\makeatother

\renewcommand\hl[1]{#1}

\begin{document}

\title{\LARGE\bf Safe Robotic Capsule Cleaning with Integrated \\ Transpupillary and Intraocular Optical Coherence Tomography}
\author{Yu-Ting Lai$^1$, Yasamin Foroutani$^1$, Aya Barzelay$^2$, and Tsu-Chin Tsao$^1$%
\thanks{This work was supported by U.S. NIH/R01EY029689 and NIH/R01EY030595.}
\thanks{$^1$ Yu-Ting Lai, Yasamin Fouroutani, and Tsu-Chin Tsao are with the Department of Mechanical and Aerospace Engineering, University of California, Los Angeles, CA, USA. 
{\tt\footnotesize \{yutingkevinlai, yforoutani, ttsao\}@ucla.edu.}}%
\thanks{$^2$ Aya Barzelay is with the Jules Stein Eye Institute, University of California, Los Angeles, CA, USA.
{\tt\footnotesize \{abarzelaywollman\}@mednet.ucla.edu.}}%
}

\maketitle

% ----------------------------------------------------------------
\begin{abstract}

\hl{Secondary cataract is one of the most common complications of vision loss due to the proliferation of residual lens materials that naturally grow on the lens capsule after cataract surgery. 
A potential treatment is capsule cleaning, a surgical procedure that requires enhanced visualization of the entire capsule and tool manipulation on the thin membrane. 
This article presents a robotic system capable of performing the capsule cleaning procedure by integrating a standard transpupillary and an intraocular optical coherence tomography probe on a surgical instrument for equatorial capsule visualization and real-time tool-to-tissue distance feedback.
Using robot precision, the developed system enables complete capsule mapping in the pupillary and equatorial regions with in-situ calibration of refractive index and fiber offset, which are still current challenges in obtaining an accurate capsule model.
To demonstrate effectiveness, the capsule mapping strategy was validated through five experimental trials on an eye phantom that showed reduced root-mean-square errors in the constructed capsule model, while the cleaning strategy was performed in three \textit{ex-vivo} pig eyes without tissue damage.}

\end{abstract}


% ----------------------------------------------------------------
\section{Introduction} \label{introduction}

% definitionTable
\begin{table*}[t!]
\centering
\caption{Abbreviations and definitions 
 of clinical and engineering terms used in this paper}
\begin{tabular}{lll}
Abbreviation & Definition & Explanation \\ \hline
PC    & Posterior Capsule                                   & The thin, transparent membrane at the back of the eye lens \\
OCT   & Optical Coherence Tomography                        & A noninvasive imaging modality utilizing low-coherence interferometry. \\ 
FK    & Forward Kinematics & A rigid transformation describing the tooltip frame with respect to the robot base frame \\
CR    & Coordinate Registration & The rigid transformation between the OCT and the robot base frame\\
\hline
\end{tabular}
\label{tab:definitionTable}
\end{table*}

\begin{figure}[t!]
    \centering
    \includegraphics[width=0.95\linewidth]{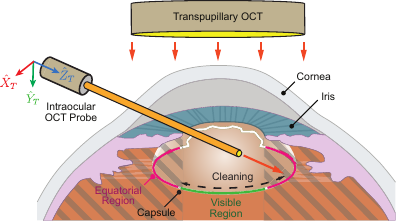}
    \caption{Illustration of the eye anatomy, transpupillary OCT, and intraocular OCT. Shaded area represents the occluded tissues that are invisible to the transpupillary OCT. The intraocular OCT is held by a robot with its tooltip coordinate system.}
    \label{fig:combined_illustration}
\end{figure}

% surgical requirements to access the equatorial region
\hl{Capsule cleaning is a potential treatment for eliminating blindness due to residual lens materials that develop around the capsular bag after cataract surgery \mbox{\cite{ucar2022posterior}}.
The procedure requires precise instrument maneuvers and timely sensing of the environment to obtain successful surgical outcomes.
Although transpupillary optical coherence tomography (OCT) and the digital microscope exhibit sufficient resolution to visualize the posterior capsule (PC) and other tissues, the shadowing effect created by the iris limits the visibility of the equatorial region and the amount of residual lens or tissue location remain unknown (Fig. {\ref{fig:combined_illustration}}) \mbox{\cite{garcia2008anterior}}.
Although polishing is theoretically feasible, many surgeons choose to skip it to avoid increased risks of capsule rupture \mbox{\cite{gonnermann2016long}}, possibly due to uncharacterized equatorial regions and inaccurate manual manipulation on the thin capsule membrane (error approximately 200--350 \textmu m) \mbox{\cite{gerber2019optical, krag2003mechanical}}.
Unlike human intervention, accurate tooltip positioning and enhanced sensing can be achieved with a robotic system that has the potential to assist and enable the polishing procedure.}

% Discuss intraocular sensing with robot and limitations
\hl{Differing from transpupillary OCT which presents limited visualization on the equatorial region, intraocular OCT probes provide access to the eye chamber that enables depth sensing of peripheral tissues intraoperatively through fiber-integrated handpieces \mbox{\cite{asami2016development, han2008handheld, abid2022development}}.
To ensure accurate tissue estimation, research and evaluations have been done using robotic arms to stabilize intraocular OCT probes that are otherwise susceptible to human hand tremor \mbox{\cite{cereda2021clinical}}.
An application is real-time model estimation on the posterior chamber for vitreoretinal procedures \mbox{\cite{cornelissen2019towards, briel2025intraoperative}}.
In both studies, 3D robotic scanning was performed with intraocular OCT measurements to predict retina boundaries, and the parameters of the spherical phantom and open-sky porcine eye models were adaptively updated.
Although the results show reliable estimates of the parameters ($<$200 \textmu m), these methods require prior knowledge of the eye position relative to the robot to generate an adequate trajectory, which cannot be established during surgery unless an additional sensor is used.
In addition, the trajectories were generated offline, but lacked real-time feedback mechanisms to account for modeling errors or tissue deformation.
}

% previous articles didn't address refractive index
\hl{Refractive index is the ratio between the speed of light in a material and the vacuum.
Different materials and shapes exhibit different optical properties that affect tissue positions measured due to OCT refractive errors \mbox{\cite{patel2019refractive, dubbelman2001shape, su2011injectable}}.
Although refractive calibration is widely used to obtain accurate tissue positions for OCT scans to reduce iatrogenic damage \mbox{\cite{stein2009adverse}}, these values are pre-calibrated prior to the procedure and may be subject to deviations such as fluid concentration change \mbox{\cite{kang2018demonstration, chen2018intraocular}}, the choice of optical fiber lens materials \mbox{\cite{mansoor2020evaluation}}, or optical path distortion \mbox{\cite{tan2022correction}}.
Furthermore, existing work often assumes that the fiber aligns with the metallic tooltip or ignores the unknown offset \mbox{\cite{zhang2009surface, song2013fiber, abid2021smart}}, which presents inaccuracies in estimating the tissue positions.
An effective method to estimate the refractive index intraoperatively and address fiber offset is important; but to the best of our knowledge, has yet to be developed for intraocular procedures.}

\hl{Intraocular procedures often rely on robot precision for tissue manipulation  \mbox{\cite{gerber2020automated}}, but we hypothesize that the cleaning procedure can be performed with reduced risks of tissue rupture with accurate tissue registration, precise capsule equator mapping from the robot, and real-time trajectory maneuvering.
Unlike other works that only rely on a single imaging modality, we propose a new strategy that achieves accuracy and precision, using both transpupillary OCT and robotic intraocular OCT to model and register the eye and visualize the occluded equatorial region, as illustrated in Fig. {\ref{fig:combined_illustration}}.

Our innovations and contributions include the following:}
\begin{enumerate}
    \item \hl{A novel capsule mapping method through a transpupillary OCT and an intraocular OCT that reduces caspule modeling errors.}
    \item \hl{The conception and realization of an online, in-situ OCT calibration workflow for the refractive index and spatial relationship during the surgery that can only be enabled with robot precision and accuracy.}
    \item \hl{A calibration approach for robot kinematics and fiber tip location that addresses the uncertainties and unknown offset from the instrument tooltip for improved tissue estimation accuracy progressively with intraocular scans.}
    \item \hl{A capsule cleaning approach on \textit{ex-vivo} pig eyes in both pupillary and equatorial regions with real-time tool-to-tissue distance feedback to ensure safety.}
\end{enumerate}

% paper organization
\hl{The rest of this article is organized as follows.
The system overview and setup is introduced in Section {\ref{sec:experimental_setup}}, followed by the preoperative steps of the procedure in Section {\ref{sec:preop_registration}}.
Section {\ref{sec:localization}} shows the technical approach for in-situ calibration and capsule mapping, and {\ref{sec:results}} demonstrates the experimental design and results for capsule mapping.
Section {\ref{sec:cleaning}} shows the capsule cleaning results on \textit{ex-vivo} pig eyes.
Finally, Section {\ref{sec:conclusion}} reviews the results and significance of the proposed methodology.
Some abbreviations and definitions of the terms used in this paper can be found in Table {\ref{tab:definitionTable}}.}

% ----------------------------------------------------------------
\section{Experimental Setup} \label{sec:experimental_setup}

\subsection{The Robotic Surgical System}

\begin{figure}[t!]
    \centering
    \includegraphics[width=0.85\linewidth]{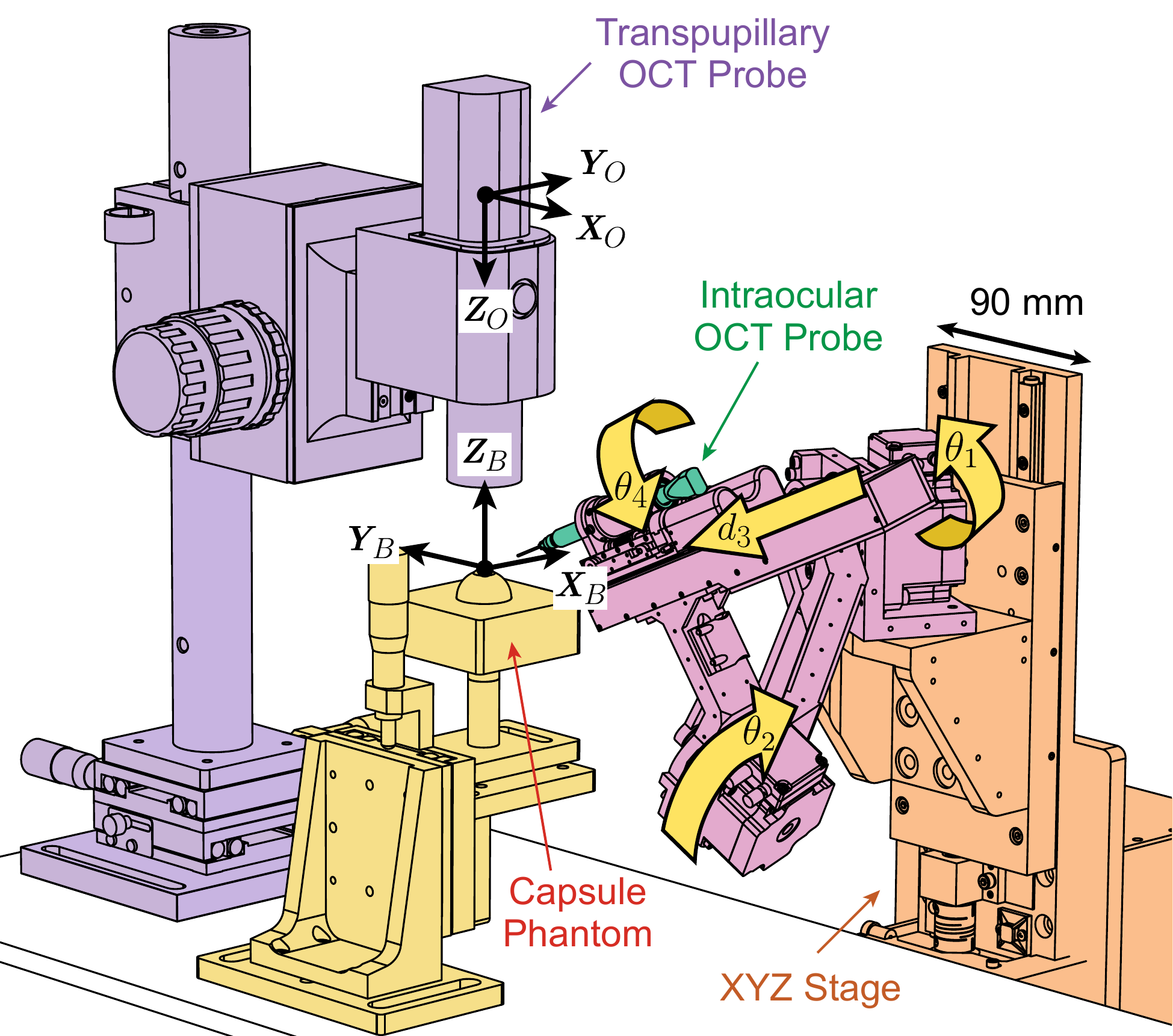}
    \caption{System overview with the definition of coordinate systems. The robot configuration shown is with $\theta_1=-50^\circ$, $\theta_2=10^\circ$, $d_3=0$ mm, and $\theta_4=0^\circ$.}
    \label{fig:robot_coordinate_system}
\end{figure}

The system architecture is illustrated in Fig. \ref{fig:robot_coordinate_system}.
\hl{A 4 degrees-of-freedom (DOF) robotic arm is used to hold the surgical instrument to ensure accurate positioning.} 
The robotic arm in our system contains a mechanically fixed remote center of motion (RCM) where the tool motion will always pass through a stationary point relative to the robot base frame $\{b\}$.
The 3D translation motion of the RCM is achieved by an XYZ stage (Optic Focus
MOXYZ-02-100-100-100), allowing the RCM to be aligned with the cornea incision by translating each joint.
Each link on the robotic arm has an incremental precision of 1 \textmu m validated through a digital gauging probe (MAGNESCALE LY-201, Sony) that has a resolution of 1 \textmu m.
A plastic probe adapter is designed and manufactured to hold the fiber in a fixed location inside a rigid metal tube

\begin{figure}[t!]
    \centering
    \includegraphics[width=0.9\linewidth]{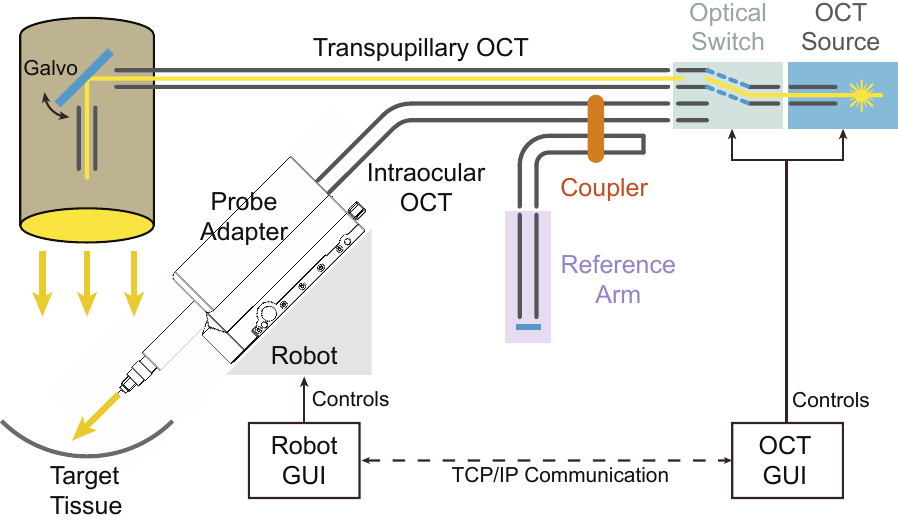}
    \caption{Shown is the illustrative OCT diagram. The robot and OCT are controlled separately through two different graphical user interfaces. The two OCT imaging shares the same source but achieves separate data acquisition with the optical switch.}
    \label{fig:systemDiagram}
\end{figure}

\subsection{OCT System}

The OCT system architecture is shown in Fig. \ref{fig:systemDiagram}. 
A spectral-domain (SD)-OCT engine with a central wavelength of 1060 nm (Telesto II, Model No. 1060LR, Thorlabs) is used for both transpupillary and intraocular OCT. 
The transpupillary OCT contains two galvanometers with an objective lens (LSM04BB, ThorLabs) that is capable of capturing three-dimensional volumetric images (V-scans) of the target tissue.
This provides a lateral resolution of 25 \textmu m and an axial resolution of 9.18 \textmu m with an imaging depth of 9.4 mm.
The OCT system is equipped with a $Z_O$-direction actuator with approximately 40 mm vertical range of motion and a resolution of approximately 1 \textmu m.
This augmentation allows for moving the vertical scanning location with respect to the eyeball for different target tissues.

To enable online acquisition of both transpupillary and intraocular OCT, an optical switch (Agiltron Inc.) is being used to perform fast switching between the two OCT modes with a latency of 5 ms.
This is considered sufficiently small compared to the acquisition rate of the transpupillary OCT (approximately 10 Hz).
The intraocular OCT contains a single optical fiber with a calibrated reference arm that can capture one-dimensional amplitude signals with respect to depth (A-scans) with a resolution of 9.4 \textmu m.
Automatically identifying the capsule with both OCT imaging modes requires prior knowledge between the two coordinate frames and precise tissue detection; therefore, calibration of robot parameters, coordinate registration, and tissue detection are characterized offline prior to experiments.

\subsection{Capsule Phantom}

\begin{figure}[t!]
    \centering   
    \includegraphics[width=0.55\linewidth]{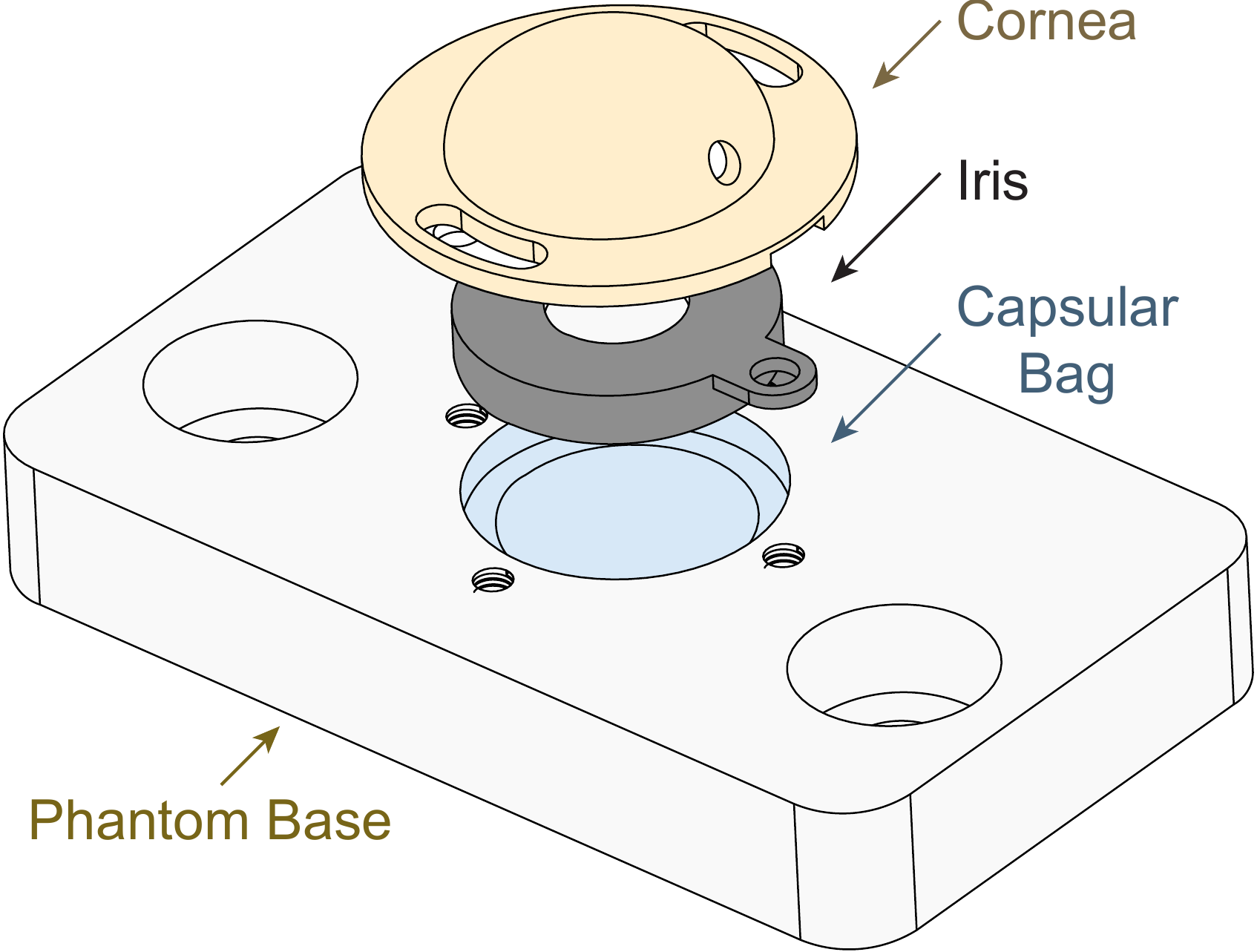}
    \caption{The eye phantom CAD model used for the mapping approach evaluation. Cornea, iris, and capsular bag can be separated to block and unblock the equatorial region.}
    \label{fig:eyePhantom}
\end{figure}

A capsule phantom was created and used to demonstrate the calibration and capsule localization results (Fig. \ref{fig:eyePhantom}).
\hl{Three sections are included in the model, including a acrylic cornea with a 2.4 mm incision, a black polypropylene iris, and a white polypropylene capsular bag.
The geometry of the equator and the lens thickness are precisely machined with a equatorial diameter of 8 mm and a 5.5 mm thickness.}
While the cornea is transparent, the iris is opaque and easily detachable from the capsular bag to block and unblock the equatorial capsule.
The eye phantom was securely placed on a stage to minimize relative motion between the transpupillary OCT, the robot, and the eye.

% ====
\section{\hl{Pre-Operative Registration and Calibration}}
\label{sec:preop_registration}

\begin{figure}[t!]
    \centering   
    \includegraphics[width=\linewidth]{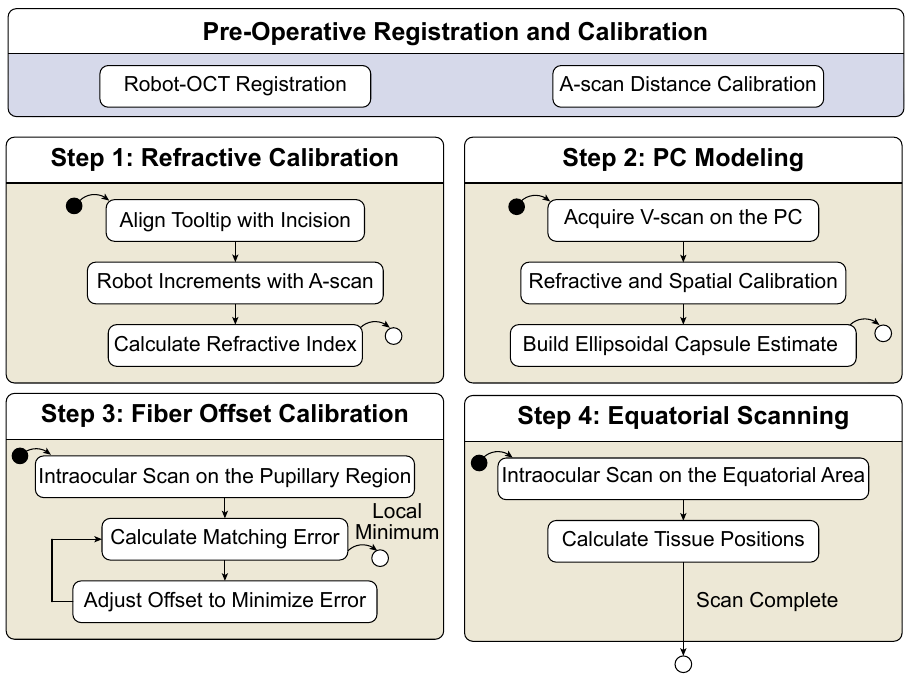}
    \caption{\hl{Shown is the schematic of the pre-operative step (offline) and the four online steps for capsule localization. Solid black dots represent the entry point of each step and white dots are the end points.}}
    \label{fig:workflow}
\end{figure}

\hl{Prior to online capsule localization, offline registration of the two coordinate systems and calibration of the A-scan distance model should be ensured.
Fig. {\ref{fig:workflow}} illustrates the steps of our proposed strategy and their respective methods are described below.}

\subsection{Robot Kinematics and Coordinate Registration}\label{subsec:manipulatorKinematics}

% \begin{figure}[t!]
%     \centering
%     \includegraphics[width=0.8\linewidth]{Figures/toolLocalization.jpg}
%     \caption{Tooltip detection result with color-coded illustration.}
%     \label{fig:toolLocalization}
% \end{figure}

%The robotic arm in our system contains a mechanically fixed remote center of motion (RCM) where the tool motion will always intersect with a stationary point relative to the robot base frame $\{b\}$.
%Each link contains an incremental precision of 1 \textmu m through a digital gauging probe (MAGNESCALE LY-201, Sony) that has a resolution of 1 \textmu m.
%A plastic probe adapter is designed and manufactured to hold the fiber in a fixed location inside a rigid metal tube, therefore the tooltip position of the intraocular probe is estimated from the Forward Kinematics (FK) of the robot.
\hl{The 4DOF robotic arm holds the A-scan fiber in place, and the tooltip position of the intraocular probe is estimated from the Forward Kinematics (FK) of the robot.}
The FK estimation of the tooltip position and orientation is derived from the transformation from robot base frame $\{b\}$ to the tooltip frame $\{t\}$:

\begin{equation}
    \prescript{b}{t}{H} = H_1H_2H_3H_4
\end{equation}
, where $H_j$ represents the homogeneous transformation from joint $\{j-1\}$ to joint $\{j\}$. 
% Standard four-parameter Denavit-Hartenberg (DH) representation with parameters $\eta_j=[d_j, \theta_j, a_j, \alpha_j]$ is sufficient to represent the first three joints, however, joint 4 requires a six-parameter representation ($\eta_4=[d_4, \theta_4, b_4, \beta_4, a_4, \alpha_4]$) to accurately estimate arbitrary tooltip position and orientation \cite{zhuang1993error, zhuang1990complete}.
% As a result, the homogeneous transformations are shown below:

% \begin{align} 
% H_j &= \textit{T}_z(d_j)\textit{R}_z(\theta_j)\textit{T}_x(a_j)\textit{R}_x(\alpha_j), \forall j\in[1,2,3] \\ 
% H_4 &= \textit{T}_z(d_4)\textit{R}_z(\theta_4)\textit{T}_y(b_4)\textit{R}_y(\beta_4)\textit{T}_x(a_4)\textit{R}_x(\alpha_4)
% \end{align}

The success of capsule localization using transpupillary and intraocular OCT relies on accurate coordinate registration (CR) between the coordinate systems of the OCT and the robot.
To establish such relationship, 20 feature points were selected within the robot workspace and the OCT field of view, and these points were used to align and calibrate the two coordinate systems.
\hl{At each point, an OCT volume scan was acquired, followed by a tooltip detection algorithm to separate the tool point cloud from the background through intensity thresholding.
Principal component analysis was performed on the tool point cloud with Gauss-Newton method to obtain a refined tool orientation.
Since the tool pose is not necessarily horizontal, the tooltip position was calculated by considering the radius of the circle fit along the tool orientation.}

The detected tooltip OCT points and robot tooltip positions from FK are then used to find the homogeneous transformation between the two sets:

\begin{equation}
    \prescript{o}{t}{H} = \prescript{o}{b}{H} \prescript{b}{t}{H}   
\end{equation}
which represents the transformation from the OCT frame $\{o\}$ to the robot tooltip frame $\{t\}$.
This is done by minimizing the following error of each feature point under $\{o\}$, where the \hl{tooltip detection} results were used as the ground truth:

\begin{equation}\label{eq:errorDefinition}
    e_i(t_{ob}) \coloneqq 
    \begin{bmatrix}
        p_{o,i} - [\prescript{o}{b}{R}(r_{ob})p_{b,i}+p_{ob}] \\
        w[z_{o,i}-\prescript{o}{b}{R}(r_{ob})z_{b,i}],
    \end{bmatrix}
\end{equation}
, where $p_{o,i}$ and $z_{o,i}$ are the  OCT tooltip position and orientation, and $p_{b,i}$ and $z_{b,i}$ are tooltip position and orientation estimated by FK parameters..
$\prescript{o}{b}{R}$ is the rotation matrix derived from $r_{ob}$.
$w$ is a predetermined weighting factor between the position and the orientation.
\hl{Through Levenberg-Marquardt (LM) optimization, the registration error between the two coordinate systems was approximately 50 \textmu m, which is in a reasonable range when accounted for OCT lateral resolution.
Although the value is still large compared to the robot precision (1 \textmu m), the tooltip positioning error is minimized to establish accurate registration between the transpupillary and intraocular OCT for capsule localization.}

\subsection{Intraocular OCT Distance Calibration} \label{sec:probeEvaluation}

\begin{figure}[t!]
    \centering   
    \includegraphics[width=0.55\linewidth]{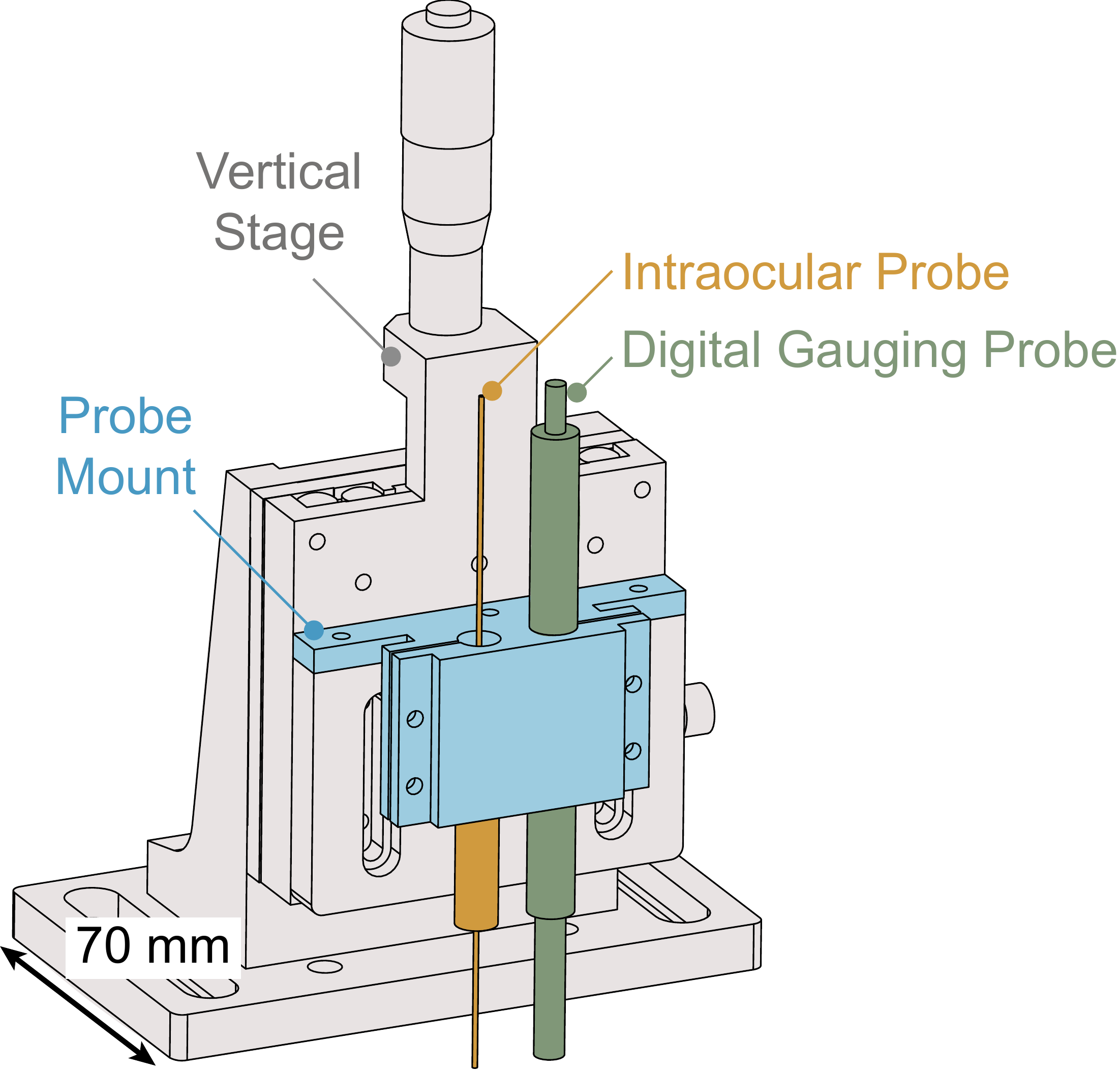}
    \caption{Shown is the setup to create the distance model and evaluate the accuracy and precision. The intraocular probe and the digital gauging probe are attached to a vertical manual stage with a resolution of 10 \textmu m. Both tips are targeting at the same surface.}
    \label{fig:ascan_sony_setup}
\end{figure}

Intraocular probe measures the reflected infrared light from the target along the fiber direction, producing an A-scan based on the distance from the fiber tip.
Higher intensities correspond to more lights reflected back from that depth.
% Fig. \ref{fig:sampleAscan} shows an example of an A-scan measurement from intraocular probe when it targets a rigid capsule surface.
A quantitative analysis is needed to compute the distance to the target surface, as intensity varies with thickness and transparency. In other words, simply choosing the highest intensity location on the graph may not represent an accurate distance, and a correct distance model is necessary. This will allow for accurate estimation of the tissue locations when combined with the robot's forward kinematics and avoids measurement inconsistencies and variations.

\begin{figure}[t!]
    \centering
    \includegraphics[width = 0.85\linewidth]{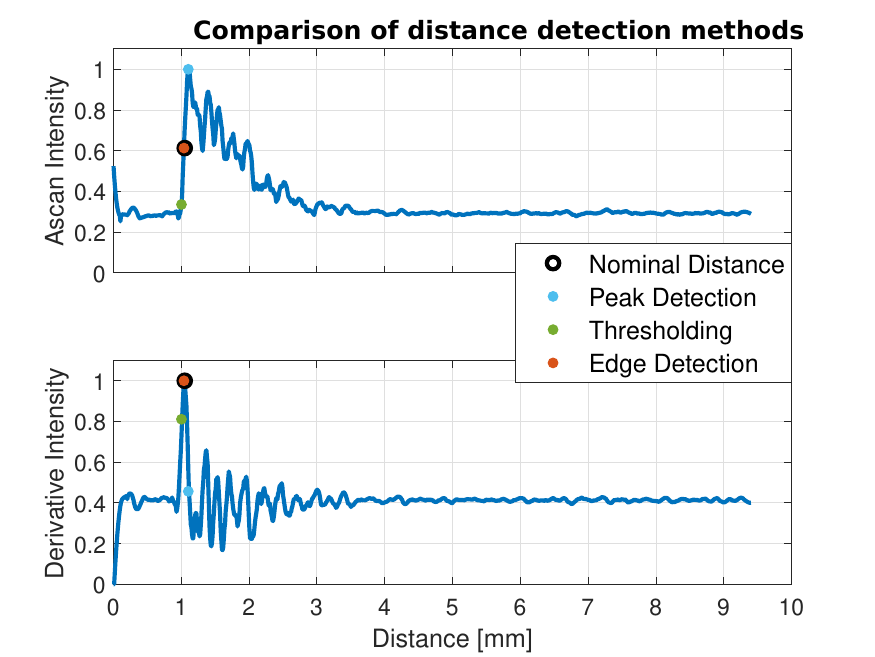}
    \caption{Results of different distance tracking models compared with the ground truth. (a) Raw A-scan signal. (b) First-order derivative of the raw A-scan signal.}
    \label{fig:sony_vs_methods}
\end{figure}

%To identify the distance detection model, the digital gauging probe was used as the ground truth reference to measure the distance to the target since it can accurately measure distances by gauging the physical displacement of the metal tube at the tip.
To identify the distance detection model, a digital gauging probe that measures distances by gauging the physical displacement of the metal tube at the tip was used as the ground truth.
As depicted in Fig \ref{fig:ascan_sony_setup}, both the intraocular OCT probe and the gauging probe were mounted in parallel on a stage that can be moved vertically with a resolution of 10 \textmu m.
A rigid, opaque, flat object was placed beneath the tips of both probes and the initial position was aligned such that both measurements started from 1 mm.
%This alignment will avoid uncertainties in the A-scan measurements when the intraocular probe is in contact with the surface and increase the robustness of distance model construction.
%Using the ground truth measurements, a distance model was created and shown in Fig \ref{fig:sony_vs_methods}.
Three detection models were analyzed: thresholding, peak detection, and edge detection.
As can be seen in Fig. \ref{fig:sony_vs_methods}, the closest correspondence with the ground truth occurs at the edge of the A-scan signal, which is the largest slope of the signal. 
% To automated the detection, a peak detection algorithm is applied on the first-order derivative of the A-scan signal for distance detection.

\begin{figure}[t!]
    \centering
    \includegraphics[width=0.75\linewidth]{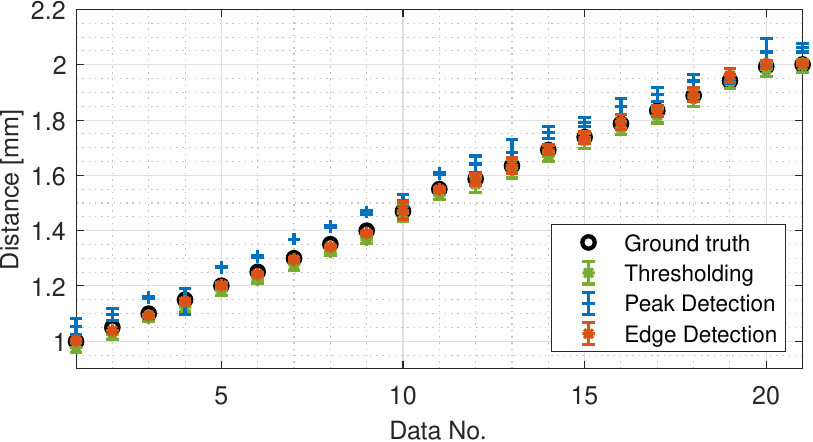}
    \caption{Distances measured by the intraocular probe and the distance gauging probe (ground truth). Different distance models are applied and compared.}
    \label{fig:sonyProbeAccuracyTest}
\end{figure}

The distance models were assessed on a set of incremental movements of 50 \textmu m performed on the vertical stage. 
%Both A-scan signal and gauging probe measurements were saved and compared at each point.
A total of 20 measurements are shown in Fig \ref{fig:sonyProbeAccuracyTest}.
By comparing the A-scan measurements with the ground truth displacement, the intraocular probe contains a root-mean-square (RMS) error of 0.33 $\pm$ 25.5 \textmu m, with a standard deviation of 9.37 \textmu m when manually labeled.
The standard deviation approximates the axial resolution of a single A-scan measurement, which is approximately 9.2 \textmu m.
This confirms that the intraocular probe is a reliable means of measuring the distance.
The measurement error statistics between different distance detection models are compared in Table \ref{tab:incrementalErrorComparison}, where the measurements from the edge detection method are more consistent with the ground truth.

\begin{table}[t!]
\caption{Measurement error of different distance model on the A-scan data}
\begin{center}
{\setlength{\extrarowheight}{1.5pt}
\begin{tabular}{c|c|c}
% \hline
Distance Model & RMS [mm]  & Std [mm]     \\ [1.5pt]
\hline
Thresholding & 0.020 & 0.008  \\ [1.5pt]
\hline
Peak Detection  & 0.055 & 0.019  \\ [1.5pt]
\hline
Edge Detection & \textbf{0.007} & \textbf{0.007}

% \hline
\end{tabular}}
\label{tab:incrementalErrorComparison}
\end{center}
\end{table}

To evaluate the accuracy and precision of the A-scan measurements, the setup as shown in Fig. \ref{fig:ascan_sony_setup} was used.
Incremental motion of 10 \textmu m was performed on the stage and $n =$ 48 data points were recorded, then the developed distance detection algorithm was applied on the A-scan signal and the distances were compared with the ground truth.
The accuracy of the measurement has a mean of 6.9 $\pm$ 20.8 \textmu m with a standard deviation of 9.7 \textmu m.
The precision was calculated by acquiring A-scan data at the same stage position 30 times, resulting in a precision of 5.6 \textmu m.
These show that the intraocular probe provides statistically accurate measurements for tissue estimation. 

\begin{figure}[t!]
    \centering
    \includegraphics[width=0.9\linewidth]{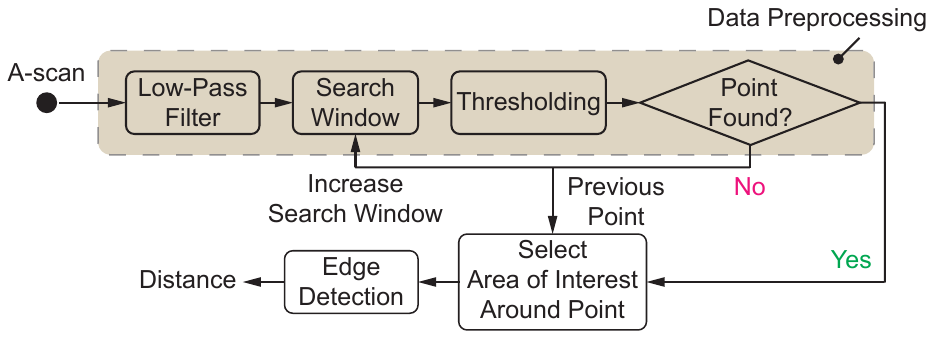}
    \caption{Flow chart for detecting the distance from A-scan data, including the pre-processing step and edge detection step.}
    \label{fig:dist_detection_flow_chart}
\end{figure}

\subsection{Distance Tracking Algorithm}\label{subsec:distance_tracking_algorithm}
% Yasamin
\hl{After determining that the peak of the derivative of the signal is the point corresponding to the A-scan distance, the detection process needs to be automated to extract the distance from raw OCT data in real time (Fig. {\ref{fig:dist_detection_flow_chart}}). 
Such an algorithm must be robust enough to handle the noise and disturbances that can occur during intraocular surgery, which were not present during the preliminary extraocular calibration process.}
Four hundred A-scan measurements were performed sequentially along the optical fiber and averaged to obtain a representative intensity signal. 
Subsequently, a zero-phase low-pass filter is applied to the acquired signal to reduce noise and artifacts, resulting in a refined signal suitable for further analysis.
As indicated in Section \ref{sec:probeEvaluation}, an edge detection model was chosen, where a first-order derivative is applied to the signal. 
However, applying a derivative decreases the signal-to-noise ratio, which poses a challenge to detecting the correct edge in real-time. 
As such, the following process is used to select the most accurate distance. 

The algorithm employs a window search approach, where an initial distance is set prior to edge detection. 
This distance serves as the center of a dynamically adjusted search window. 
Thresholding is then applied within this window to identify an estimate of the distance. 
Subsequently, a new window is created around this estimate to search for the signal edge. 
If the edge is detected, it becomes the current distance; otherwise, the previous estimate is retained. 
If no point is found above the desired threshold, the algorithm reverts to the distance from the previous step and increases the search window width for the next iteration. 
Whenever a new point is successfully identified by the algorithm, the window size is reset.

The search window ensures the correct selection of the desired edge, preventing other intraocular tissues near the probe from being mistaken for the target tissue. This method assumes minimal sudden movements between the robot and tissue, so the new distance is expected to vary only slightly compared to the previous value, remaining within the search range.
The threshold is determined based on the average signal intensity in the last 20\% of the signal, which is considered noise because it falls outside the working distance of the A-scan probe. 
This average value is multiplied by an empirically determined gain to establish a threshold for the distance signal. 
The gain can be modified if there is a significant change in the signal-to-noise ratio.

% ----------------------------------------------------------------
\section{\hl{Technical Approach to Capsule Localization}}
\label{sec:localization}

\hl{The online capsule localization approach is segmented into four steps.}
\begin{enumerate}
    \item \hl{Refractive calibration: The robot moves in small increments and online calculates the refractive index associated with A-scan distances.}
    \item \hl{PC modeling: Transpupillary OCT takes a V-scan to model the posterior capsule and register eye location.}
    \item \hl{Fiber offset calibration: To eliminate the error created by uncertain fiber location.}
    \item \hl{Equatorial scanning: The intraocular is commanded to "look at" the peripheral region.}
\end{enumerate}

\hl{These steps are done in-situ and are detailed in this section.}

\subsection{Step 1: Refractive Calibration}\label{subsec:opticalCalibration}

\begin{table}[t!]
    \caption{Results of refractive index calibration on different mediums}
    \begin{center}
    {\setlength{\extrarowheight}{1.5pt}
    \begin{tabular}{c|c|c|c}
        Material & Calibration Result & Nominal Value & Error \\
        \hline
         Air & 1.009 & 1 & 0.9\% \\
         Gel & 1.353 & 1.384 & 2.24\% \\
         Water & 1.332 & 1.333 & 0.07\% \\
         BSS & 1.329 & 1.3345 & 0.42\% \\
         Lens Material & 1.38 & 1.40 & 1.40\% \\
         Vitreous & 1.332 & 1.3345 - 1.336 & 0.20\%\\
    \end{tabular}

    \label{tab:refractive_index}}
    \end{center}
\end{table}

Accurate distance detection with the intraocular probe requires adjusting for the refractive index of the medium through which the beam is passing. The refractive index for each medium is often assumed to be fixed, but fluid exchange for intraocular procedures can create different medium mixtures.
This generates different optical distortion of the intraocular measurements between experiments, altering the estimation of the physical distance of the intraocular tissues.
The optical calibration utilizes the precision of the robot where a 1 \textmu m incremental motion on $d_3$ can be achieved and the movement is invariant to the intraocular environment.
To perform calibration, the intraocular probe first aims at a target tissue, and an incremental motion of $d_3=50$ \textmu m is performed with a total of 20 data points. 
During each increment, the developed distance tracking algorithm is applied to the A-scan data and the final converged distance is recorded.
The slope of these distances over the $d_3$ encoder readings is selected as the refractive index of the medium being analyzed. 

To evaluate the accuracy of the calibration method, the values are compared with the nominal values presented in \cite{stupar2012remote, uhlhorn2008refractive, mishra2023vitreous}.
Five different mediums are analyzed, including air, BSS, viscoelastic, lens material, and vitreous.
BSS and viscoelastic are the medium that are often used in surgery whereas lens material and vitreous naturally exist in human eyes. 
As shown in Table \ref{tab:refractive_index}, the calibrated results for each medium generally have less than 1\% difference from the nominal values, except for a difference of 2.24\% in gel.
This shows the effectiveness of the optical distortion calibration method and its real-time application in intraocular procedures.

\subsection{Step 2: PC Modeling and Registration}\label{subsec:spatialCalibration}

\begin{figure}[t!]
    \centering
    \includegraphics[width=0.9\linewidth]{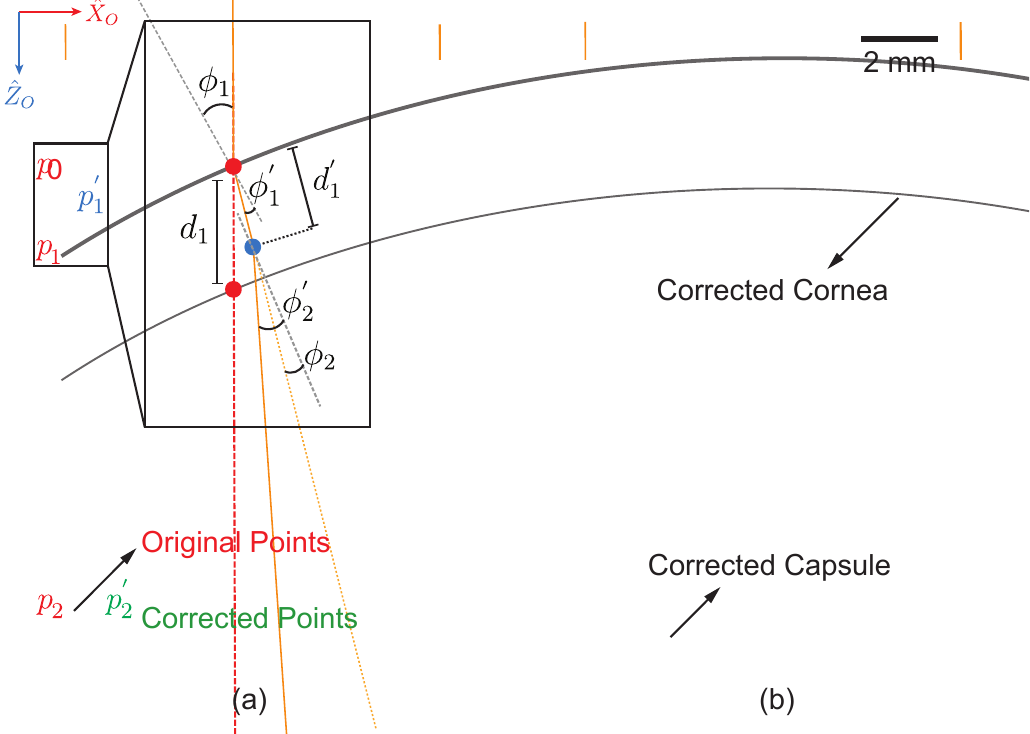}
    \caption{Illustration of the spatial calibration of the cornea and PC for each A-scan and resulting curves. (a) Uncalibrated OCT scan. (b) Calibrated OCT scan. Original points are marked as red and the calibrated points are marked as green and blue. The data is extracted from an actual B-scan image.}
    \label{fig:bscan_zoomed_in}
\end{figure}

\begin{figure}[t!]
    \centering   
    \includegraphics[width=0.7\linewidth]{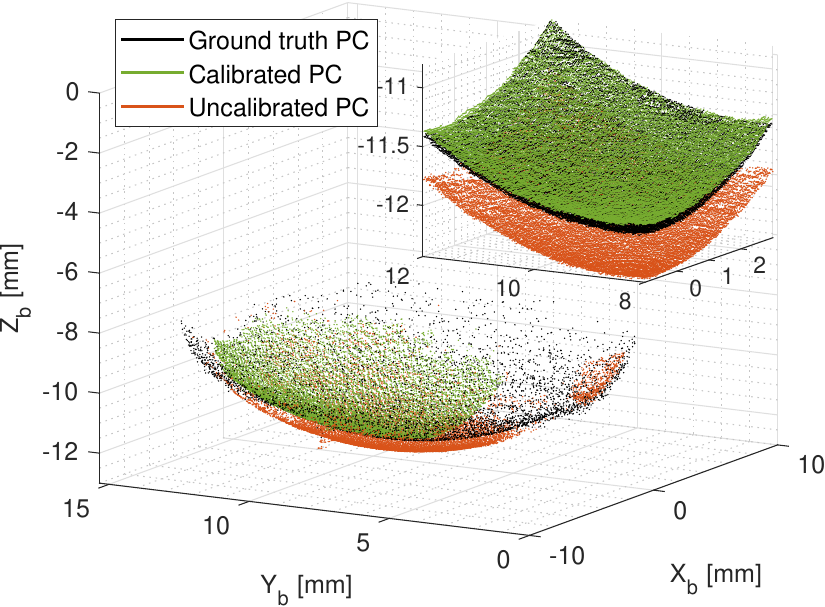}
    \caption{3D visualization of the PC with and without calibrating spatial distortion in the transpupillary V-scan due to cornea geometry.}
    \label{fig:bscanCalibrationVisualization}
\end{figure}

An OCT volume scan of the visible capsule was acquired to register the eye location through the established registration and model the capsule anatomy for intraocular scanning.
Due to the natural convex shape of the cornea, the infrared light emitted from the OCT system introduces spatial distortion due to Fermat's principle in Eq. (\ref{eq:fermats_principle}) when it travels between two different mediums.
\begin{equation} \label{eq:fermats_principle}
    n_1d_1 = n_2d_1'
\end{equation}
where $n_1$ and $n_2$ are the refractive indices of the medium, $d_1$ is the undistorted distance traveled, and $d_1'$ is the actual distance after distortion due to the medium change. 
Spatial distortion occurs at intermediate tissue boundaries, where the incidence angles vary for each A-scan due to the curvature of the cornea, resulting in inaccurate tissue positions with respect to $\{o\}$.
The angle of refraction ($\phi_1^{'}$) can be calculated from Snell's law:
\begin{equation} \label{eq:snells_law}
    n_1 \sin(\phi_1) = n_2 \sin(\phi_1'),
\end{equation}
where $\phi_1$ is the angle of incidence. 
This can be used to correct for the spatial distortion of each B-scan image.

Fig. \ref{fig:bscan_zoomed_in} illustrates the distortion calibration for the anterior chamber, highlighting both the original and corrected tissue locations.
Similarly to the method presented in \cite{tan2022correction}, a segment-by-segment calibration method was implemented to consider intermediate tissues along the light path. 
Through segmentation of each B-scan image, it is possible to detect the outer and inner boundaries of the cornea, as well as the posterior capsule. 
An ellipse was fitted to the cornea edges and the slope of the tangent line was used to calculate the angle of incidence ($\phi_1$) of each A-scan entering the cornea. 
Known refractive indices were assumed for the cornea (1.38 for human eyes \cite{patel1995refractive} and 1.45 for acrylic), and the inner cornea can be corrected by combining Eq. (\ref{eq:fermats_principle}) and (\ref{eq:snells_law}).

Following a similar process, the corrected inner cornea was used to calculate the incidence angle of the beam entering the anterior chamber. 
Thus, the refraction angle and corrected PC points can be derived from their measured geometry.
This step requires knowledge of the refractive index of the medium inside the eye ($n_3$), which was identified in Sec. \ref{subsec:opticalCalibration}.
This procedure is applied on each optical path to reconstruct a calibrated PC point cloud with the following depth correction:

\begin{equation}
   z_0'=
   \begin{cases}
     z_0, &  z_0 < z_e \\
     z_e + n(z_0-z_e), & z_0 \geq z_e
   \end{cases}
\end{equation}
where $z_0$ is the initial tissue location, $z_0'$ is the adjusted depth value under the OCT image, $z_e$ is the outer surface of the cornea and $n$ is the ratio of the refractive index of the air over that of the medium.

\begin{figure}[t!]
    \centering   
    \includegraphics[width=0.75\linewidth]{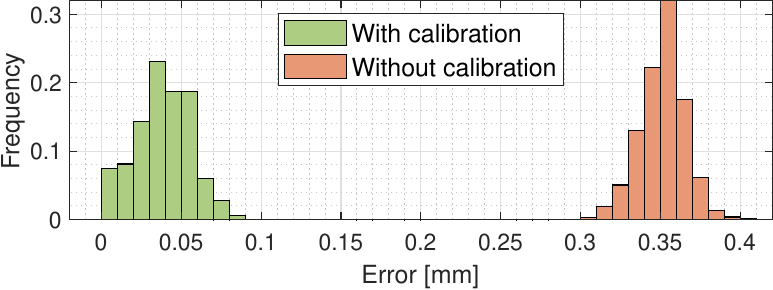}
    \caption{Error histogram comparison with and without spatial calibration.}
    \label{fig:bscanCalibrationHistogram}
\end{figure}

The results are compared with the ground truth PC point cloud, which was acquired when the cornea was removed with no medium present in the chamber.
The uncalibrated and calibrated capsule are visualized in Fig. \ref{fig:bscanCalibrationVisualization}, where the calibrated PC aligns well with the ground truth and achieves a significantly reduced RMS error (Fig. \ref{fig:bscanCalibrationHistogram}).
As reported in Table \ref{tab:spatialCalibError}, spatial calibration minimizes the mean error between measurement and ground truth data but with a similar standard deviation.
With the PC point cloud corrected, an ellipsoidal model fit was generated with prior knowledge of the pupil location.

%A guide to the mathematical notations used is also presented in Table \ref{tab:bscanCalibrationSymbols}

% \begin{table}[t!]
%     \centering
%     \caption{Mathematical notations for Spatial Calibration}
%     \label{tab:bscanCalibrationSymbols}
%     {\setlength{\extrarowheight}{1.5pt}
%     \begin{tabular}{c|l}
%         Symbol & Description \\
%         \hline
%          $d_i$ & Perceived distance in layer $i$ \\
%          $d_i'$ & Calibrated distance in layer $i$\\
%          \multirow{2}{*}{$\phi_i$} & Incidence angle entering layer $i$ \\
%          & with respect to the surface normal\\
%          \multirow{2}{*}{$\phi_i^{'}$} & Refraction angle going through layer $i$ \\
%          & with respect to the surface normal\\
%          $\gamma_i$ & Incidence angle relative to the $\hat{Z}_O$ axis (vertical line)\\
%          $\gamma_i'$ & Refraction angle relative to the $\hat{Z}_O$ axis\\
%          $n_i$ & Refractive index of layer $i$\\
%          $s_i$ & Slope at the incidence point of layer $i$\\
%          $p_i$ & Each perceived point in layer $i$\\
%          $p_i'$ & Corrected points in layer $i$\\
%          $i = 1$ & Cornea layer\\
%          $i = 2$ & Anterior chamber layer\\
%          \end{tabular}}
% \end{table}

\begin{table}[t!]
\caption{Spatial Calibration Error}
\begin{center}
{\setlength{\extrarowheight}{1.5pt}
\begin{tabular}{c|c|c}
% \hline
Error Type   & RMS [mm]  & Std [mm] \\ [1.5pt]
\hline
Before Spatial Calibration & 0.352 & 0.014 \\ [1.5pt]
\hline
After Spatial Calibration & 0.042 & 0.017 \\
% \hline
\end{tabular}}
\label{tab:spatialCalibError}
\end{center}
\end{table}

%Through a similar process as \cite{tan2022correction}, the effective refractive index of transpupillary OCT scans inside the eye can be calculated. This is necessary because the medium inside the eye can change between trials and even throughout the experiment, and the images need to be adjusted accordingly to give the most accurate modeling of the eye.  To do so, scanning is done along the tool axis when the tool is inside the eye. The robot moves in 0.1 mm increments along the tool axis, and the tooltip is detected through traditional image processing by binarizing the image and searching over the range of interest. The detected depth in the Bscans is then compared to the known depth of the robot tooltip, and the ratio of the slopes of the best fit line can be considered as the effective refractive index of the eye. We specify that this is the \textit{effective} refractive index, because it combines all the medium in the OCT view, including cornea and lens material if applicable, and calculates the overall approximation of the value that would correct for the final tooltip location.

\subsection{Step 3: Fiber Offset Calibration}

Once the error due to refractive and spatial distortion are minimized, the remaining error is the uncertain fiber offset inside the metallic tool ($d_{tip}$) and any misalignment may result in inaccurate tissue estimation from the tooltip.
The goal is to acquire the two measurements on the same reference tissue and minimize the error associated with the $d_{tip}$.
To accomplish this, the corrected PC point cloud was used as the reference tissue, followed by image processing to get the ground truth point cloud of the surface of interest.
The robotic arm was commanded to move to 20 points generated within the pupillary region with A-scan measurements acquired at each robot position.
These A-scan measurements were used to estimate the target tissue positions by extending the tooltip positions along tooltip orientation with the distances estimated by the A-scan, assuming the registration is accurate from Section \ref{sec:preop_registration}.
Each estimated tissue from the intracoular OCT fiber $p_b^c$ can be described by the equation below.
\begin{equation}
    p_b^c = p_b^t + z_b^t \cdot d
    \label{eq:tissueEst}
\end{equation}
where $p_b^t$ and $z_b^t$ are tooltip position and orientation calculated from the calibrated registration, and $d$ is the distance estimated from the A-scan measurements.
A local fast nearest neighbor search algorithm was performed to find the transpupillary points ($p_o^c$) that are closest to the estimated $p_b^c$. 
The error metric was defined as the average Euclidean distance between them and $d_{tip}$ can then be found by minimizing the following objective function through gradient descent.
\begin{equation}
    \min_d \frac{\sum_N \|p_b^c - p_o^c\|_2}{N}
    \label{eq:minimize_dtip}
\end{equation}

\begin{figure}[t!]
    \centering
    \includegraphics[width=0.7\linewidth]{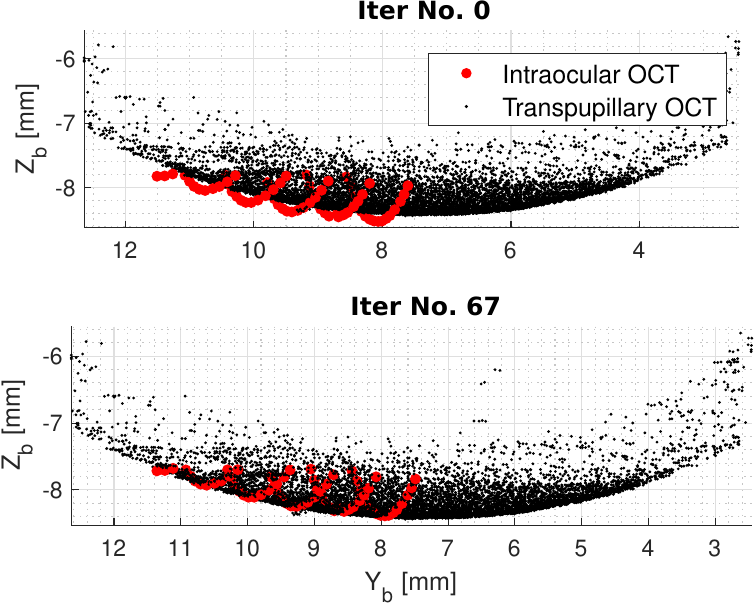}
    \caption{Shown is the side view of the projection before (Iter No. 0) and after (Iter No. 67) fiber offset calibration. The project error between the intraocular OCT and transpupillary OCT is minimized in the bottom plot.}
\label{fig:probeCalibration}
\end{figure}

\begin{table}[t!]
\caption{Probe Calibration Error}
\begin{center}
{\setlength{\extrarowheight}{1.5pt}
\begin{tabular}{c|c|c}
% \hline
error type   & RMS [mm]  & Std [mm] \\ [1.5pt]
\hline
Before Probe Calibration & 0.352 & 0.130 \\ [1.5pt]
\hline
After Probe Calibration & 0.074 & 0.030 \\
% \hline
\end{tabular}}
\label{tab:CRError}
\end{center}
\end{table}

\begin{figure}[t!]
\centering
\includegraphics[width=0.85\linewidth]{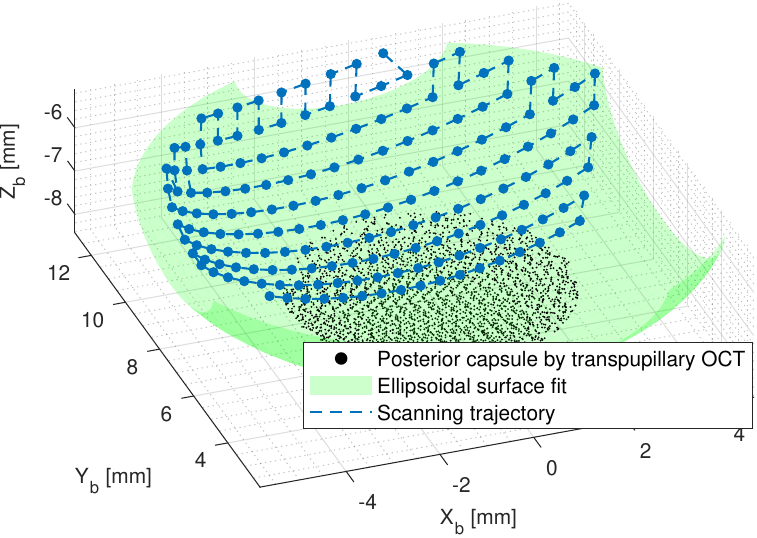}
\caption{Tooltip scanning trajectory generated by traveling salesman problem with an nominal ellipsoidal modeling on the transpupillary OCT data.}
\label{fig:tsp_results}
\end{figure}

\subsection{Step 4: Intraocular OCT Scanning Trajectory}

Following the calibration procedures, an intraocular scanning trajectory was generated based on the ellipsoidal fit of the capsule and within the robot workspace.
The generated model is not accurate because only pupillary capsule data was considered in the ellipsoidal fit, causing inaccuracy near the equatorial region.
Therefore, a trajectory offset of 1 mm along the tool direction and away from the ellipsoid was employed, which compared to other calibration and registration error, is sufficiently large to account for the modeling error.
Fig. \ref{fig:tsp_results} shows the posterior capsule acquired by the transpupillary OCT and the ellipsoidal surface fit model from the data with least squares fit.

Assuming the capsular bag of a human eye is convex and continuous \cite{smith2009mathematical}, exhausted and high-resolution scanning is not required for the tooltip.
As a result, a set of $\theta_1$ and $\theta_2$ robot angles with 3$^\circ$ increments were selected programmatically that are within the constraints of the robot workspace and the boundary of the ellipsoidal fit.
These points were mapped to the OCT frame through established coordinate transformation, and projected to the ellipsoidal fit with an 1 mm offset from the surface as the waypoints of the trajectory.

To efficiently travel through the waypoints and reduce travel distance, a Traveling Salesman Problem (TSP) is formulated to solve the optimal traveling strategy through linear programming.
The objective is to minimize robot joint motion throughout the travel and can be described as the weighted absolute travel distance of the joints between points $i$ and $j$: 
\begin{equation}\label{eq:weighted_distance}
    J = k \begin{bmatrix}
        \Delta \theta_1 \\ \Delta \theta_2 \\ \Delta d_3
    \end{bmatrix}_{ij},
\end{equation}
where $k \in \mathcal{R}^{1\times3}$ is the weighting matrix.
Due to model uncertainties and motor backlash, the weighting factor for $\Delta d_3$ was selected to be ten times larger to penalize large movements along the tool shaft. %, and $k$ is selected to be $[1, 1, 10]$.
The optimization will generate a trajectory where each waypoint is traveled once and only has two associated trips (arrival and departure). 
The optimized trajectory is illustrated in Fig. \ref{fig:tsp_results}, where the dots are the generated waypoints.
\hl{The distance tracking algorithm was applied during tooltip tracking and the A-scan data was acquired at each waypoint.}

\section{Results on Capsule Localization} \label{sec:results}

This section demonstrates capsule cleaning on \textit{ex-vivo} pig eyes using the capsule mapping approach and intraocular OCT A-scan feedback to ensure safe distance of the surgical instrument from the capsule.

\subsection{Experimental Design}
\hl{To comprehensively evaluate the effectiveness of our proposed process, each step is performed under different conditions and compared with existing strategies. 
First, online localization is performed in air, then repeated with gel added to simulate the lens or vitreous of the eye. 
The results are compared with non-calibrated data in the presence of gel, highlighting the impact of in-situ calibration in accounting for various error sources.
Next, intraocular scanning is performed using a dense set of A-scan measurements (approximately 180 data points), followed by a systematic reduction in point density. 
This allows us to assess modeling error at different densities and determine the optimal set of waypoints for precise modeling.
Finally, after scanning, PC modeling is conducted. Initially, it is done using only transpulillary OCT data visible through the iris. 
The model is then expanded by incorporating progressively larger portions of the A-scan measurements of the periphery, and the fitting error is evaluated against ground truth measurements.}
% Fig. \ref{fig:capsuleLocalizationWithTSP} further illustrates the relationship described in (\ref{eq:tissueEst}). 

% \begin{figure}[t!]
% \centering
% \includegraphics[width=0.85\linewidth]{Figures/capsuleLocalizationWithTSP.pdf}
% \caption{Illustration of (\ref{eq:tissueEst}). Example trajectory was downsampled to demonstrate the relationship. Blue dots represent the designed trajectory ($p_b^t$), while the purple dots are the corresponding capsule positions ($p_b^c$) estimated from $d$ along the tooltip orientation ($z_b^t$).}
% \label{fig:capsuleLocalizationWithTSP}
% \end{figure}

\subsection{Capsule Mapping Results}

\begin{figure}[t!]
\centering
\includegraphics[width=0.95\linewidth]{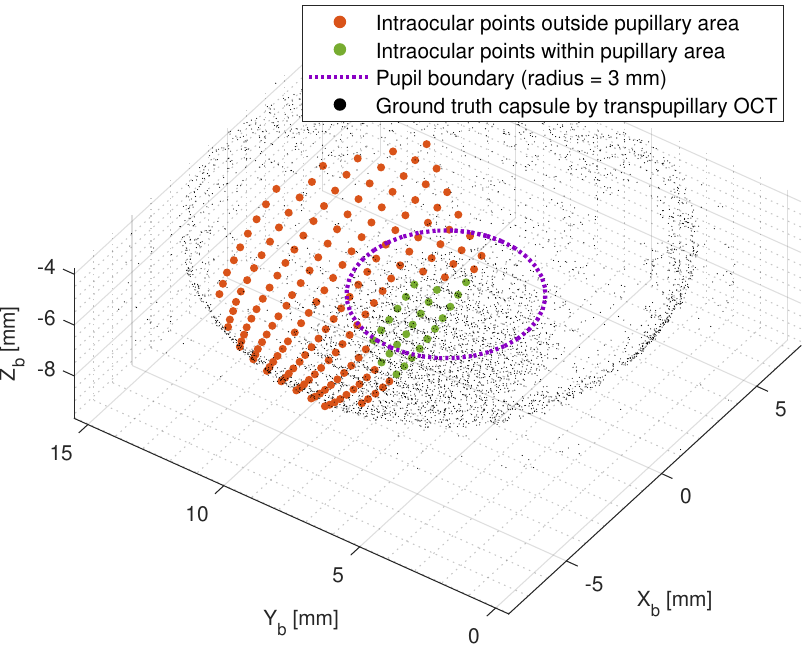}
\caption{Shown is the result of capsule mapping results. The black points are the ground truth generated by the transpupillary OCT when the iris is removed. The intraocular OCT generates the points within the pupillary area (green dots) and visualizes the equatorial area (orange dots).}
\label{fig:mapping_result}
\end{figure}

\begin{figure}[t!]
\centering
\includegraphics[width=0.55\linewidth]{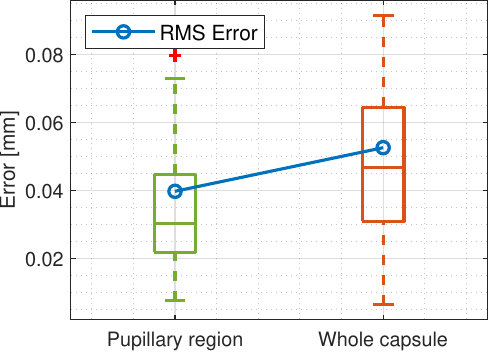}
\caption{\hl{Localization error statistics between the pupillary region and equatorial region. The RMS error is similar to robot-OCT registration error but presents slight increase with the addition of equatorial region due to sparser ground truth data.}}
\label{fig:localization_error_air}
\end{figure}

\subsubsection{Equatorial Capsule Localization Accuracy}

\hl{To show the effectiveness of the developed capsule localization method, the identified $p_b^c$ is numerically compared with the ground truth volume scan from the transpupillary OCT.
A volume scan is first taken with the cornea and iris, followed by thresholding on the volume scan to estimate the ground truth capsule location ($p_{g}^c$).
Then a scanning trajectory is generated using the visible posterior capsule data.
After completing equatorial scan and capsule localization, the cornea and iris are manually removed to establish the ground truth by a transpupillary OCT V-scan for comparison.
As shown in Fig. {\ref{fig:mapping_result}}, $p_b^c$ covers points that are within the pupillary area (green dots) and peripheral region (orange dots), while the transpupillary OCT can only see the data points within the pupillary area (purple circle).}

\hl{The localization error is defined as the Euclidean distance between $p_b^c$ and the closest point in the volume scan to $p_b^c$:}
\begin{equation}
    e_i = \|p_{b,i}^c - p_{g,k}^c\|_2,
\end{equation}
\hl{where $p_{g,k}^c$ represents the ground truth capsule data $k$ that is closest to $p_b^c$ from a local fast nearest neighbor search.
To efficiently find $p_{g,k}^c$ on the dense ground truth data, only 200 \textmu m cubic area centered at $p_b^c$ is used in the V-scan.
The statistics between 5 trials (Fig. {\ref{fig:localization_error_air}}) showed that the intraocular OCT has an RMS error of 39.8 \textmu in estimating the pupillary capsule and an RMS error of 52.6 \textmu m for the whole capsule.
The measured accuracies are close to the robot-OCT registration error (approximately 50 \textmu m), which suggests that the major source of error subjects to OCT lateral resolution.
However, the reported RMS error increases with more equatorial points may be due to sparser ground truth data in the OCT V-scan at periphery when the surface is almost parallel to the incident light.}

\begin{table}[t!]
\caption{Capsule localization error with the anterior chamber filled with gel}
\begin{center}
{\setlength{\extrarowheight}{1.5pt}
\begin{tabular}{c|c|c|c}
% \hline
Method & Error Region & RMS [mm]  & Std [mm] \\ [1.5pt]
\hline
\multirow{ 2}{*}{Without Calibration} & Pupillary & 0.796 & 0.393 \\ [1.5pt]
& Whole Capsule & 0.938 & 0.470 \\
\hline
\multirow{ 2}{*}{With In-Situ Calibration} & Pupillary & 0.067 & 0.031 \\ [1.5pt]
 & Whole Capsule & 0.057 & 0.028 \\
\end{tabular}}
\label{tab:localizationError}
\end{center}
\end{table}

\subsubsection{Localization with Gel-Filled Chamber}

\hl{The anterior chamber is filled with gel to evaluate the performance and the effectiveness of in-situ calibration under viscoelastic conditions.
Localization results with and without in-situ calibration are compared and the reported RMS error of $p_b^c$ in the pupillary region and the whole capsule are shown in Table {\ref{tab:localizationError}}.
When the anterior chamber is filled with gel, in-situ calibration becomes critical due to significant spatial distortion of the OCT V-scan at periphery.
The error was substantial without in-situ calibration, which may be unsuitable for intraocular procedures especially at the equatorial region that cannot be visualized by the transpupillary OCT.
Our approach is capable of reducing the RMS error of the whole capsule from 938 \textmu m to 57 \textmu m.
This reduces modeling uncertainties and potentially provides safer tooltip targeting in the chamber without interfering with the tissues.}

% \subsubsection{Improving Capsule Modeling}

% \hl{Initially the capsule modeling only relies on the transpupillary data, which had a estimation error of approximately 4\% in RMS error in the capsule radii.
% We demonstrate an iterative update of the capsule model as more intraocular points are scanned and added to the model fitting (Fig. {\ref{fig:error_progression}}).
% An increment of 20 points was added progressively to estimate the capsule radii in the ellipsoidal model, and the final error is less than 0.5\% from the designed value.}

% \begin{figure}[t!]
% \centering
% \includegraphics[width=0.85\linewidth]{Figures/error_progression.pdf}
% \caption{\hl{Shown is the estimation error progression with the addition of intraocular measurements. The estimated error of the ellipsoidal model reduces with more intraocular points are considered.}}
% \label{fig:error_progression}
% \end{figure}

\subsubsection{Scanning Trajectory}

\begin{figure}[t!]
\centering
\includegraphics[width=0.9\linewidth]{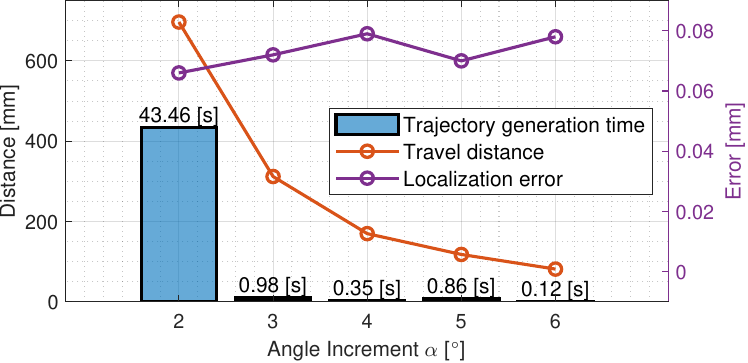}
\caption{\hl{Shown is the trajectory generation time and error with respect to various joint angle increments ($\alpha^\circ$) for trajectory generation. The optimization for 2$^\circ$ increments presents large delays and travel distance compared to other angles.}}
\label{fig:tsp_time_distance}
\end{figure}

\hl{To identify an adequate joint angle increment ($\alpha$) for trajectory generation, capsule localization was conducted for various angle increments $\alpha$, and the corresponding elapsed time, travel distance, and localization error were compared in Fig. {\ref{fig:tsp_time_distance}}.
Although $\alpha=2^\circ$ has the smallest localization error, the trajectory generation time is significantly larger than other joint increments during scanning trajectory optimization.
With larger scanning increments, the optimization process did not produce large delays in the whole procedure but slightly increased the localization error.
Given the travel distance, the elapsed time, and the localization error, $\alpha=3^\circ$ was selected in this study, but more analysis on biological models with different variability should be conducted to prove its clinical advantages.}

\section{Capsule Cleaning on Animal Models} \label{sec:cleaning}

\subsection{Experimental Workflow and Pig Eye Preparation}

\begin{figure}[t!]
    \centering
    \includegraphics[width=0.95\linewidth]{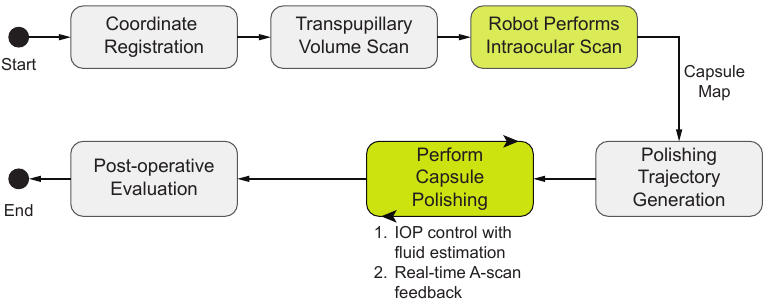}
    \caption{Proposed automated workflow for capsule polishing on a pig eye.}
    \label{fig:cleaning_workflow}
\end{figure}

The experimental workflow is illustrated in Fig. \ref{fig:cleaning_workflow}.
Coordinate registration and robot calibration were performed offline prior to the experiments.
Once registration is complete, the capsule mapping procedure with transpupillary and intraocular OCT is performed in the pig eye to construct a complete and updated capsule map.
Then the cleaning trajectory is generated within the capsule map to perform the procedure.
During cleaning, active intraocular pressure control is turned on (the same system in \cite{lai2024add}) and real-time A-scan feedback is enabled to adaptively update the $\theta_3$ angle and maintain a fixed safe distance from the capsule.

\begin{figure}[t!]
\centering
\includegraphics[width=\linewidth]{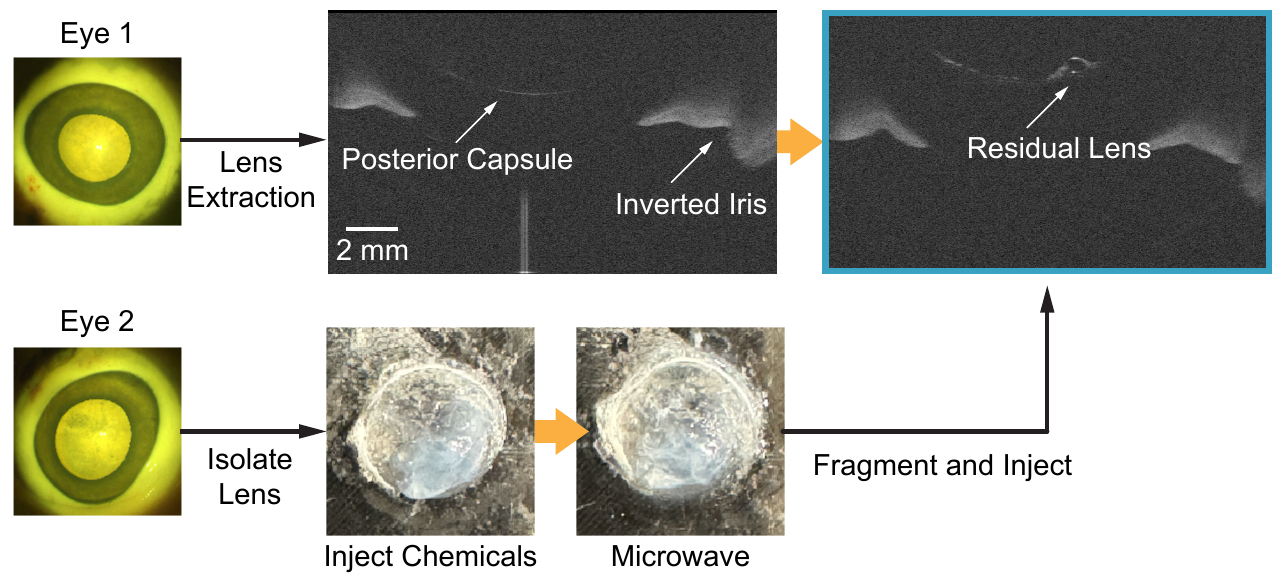}
\caption{Shown is the steps to create residual lens materials on top of the cleaned capsule. The first eye undergoes standard manual lens removal process and the lens in the second eye was fully taken out, induced with cataract, and injected back to the anterior chamber of the first eye.}
\label{fig:residual_lens_creation}
\end{figure}

Different methods have been explored to produce residual lens materials and increase the contrast of the capsule \cite{gerber2021robotic, lee2023accurate}, but the fluid properties of the generated residual lens are not consistent and difficult to control.
The approach to creating the residual lens requires two pig eyes.
The lens of the first eye was completely removed with an intact capsular bag remained as an empty container.
The lens of the second eye was isolated, hardened with a chemical solution and microwave \cite{shentu2009combined}, fragmented, and injected back into the first eye to simulate realistic residual lens materials on the capsule.
Finally, a sterile lubricating jelly (MDS032290H, Medline) was injected to fill the anterior chamber to prevent structural collapse and maintain the shape of the capsular bag.
The resultant residual lens is shown in Fig. \ref{fig:residual_lens_creation}.

\subsection{Imaging of Intraocular Tissues}

\begin{figure}[t!]
    \centering   
    \includegraphics[width=\linewidth]{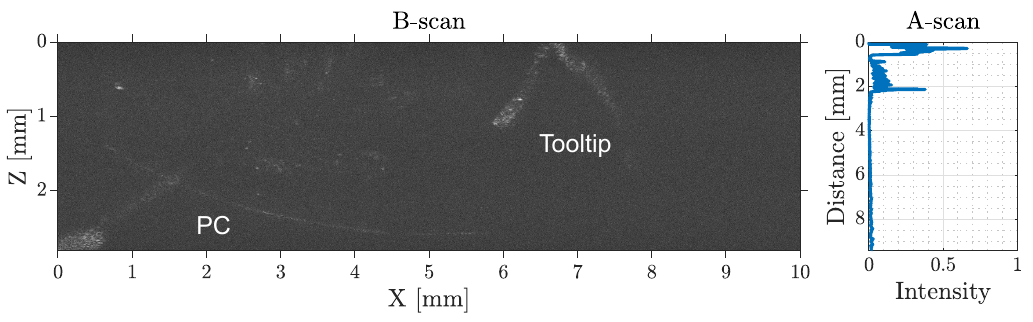}
    \caption{Shown is the cross-validation of intraocular tissue measurements between A-scan and B-scan.}
    \label{fig:merge_tissue_ascans}
\end{figure}

Prior to the experiment, it is necessary to ensure that intraocular OCT creates valid distance measurements of the target tissue in an enclosed intraocular environment.
The tooltip was placed close to and aimed at the lens capsule, while the transpupillary OCT scan was positioned so that it scanned along the tool shaft and covered the tool and tissue simultaneously.
Once the focus of both OCT measurements was properly adjusted, the distances between the tooltip and tissue were compared.
Fig. \ref{fig:merge_tissue_ascans} shows the OCT B-scan of the tool and the tissue and its corresponding A-scan measurements, indicating that the intraocular OCT probe is capable of visualizing target tissues in intraocular environments.

\subsection{Anatomical Modeling and Capsule Mapping}

To construct the capsule map, a scanning trajectory is first generated for robotic intraocular scans.
We used transpupillary OCT volume scans to visualize and model the cornea, iris, and posterior capsule.
A deep learning segmentation model was used to obtain 3D point clouds for the cornea and iris, followed by pupil detection to obtain the center and orientation of the eye.
The capsule model was generated using thresholding and image processing to obtain the location of the posterior capsule, update the ellipsoid parameters, and obtain a more accurate capsule estimate to generate an intraocular scanning trajectory. 

\begin{figure}[t!]
    \centering   
    \includegraphics[width=0.85\linewidth]{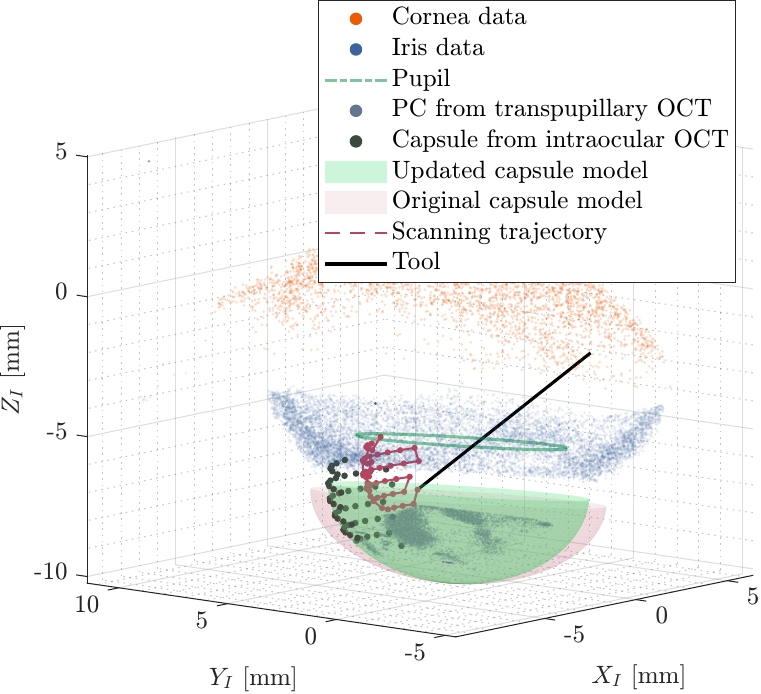}
    \caption{Shown is the anatomical modeling results.}
    \label{fig:modeling_results_ellipsoid}
\end{figure}

Fig. \ref{fig:modeling_results_ellipsoid} shows the critical intraocular tissues (cornea, iris, posterior capsule) with capsule modeling and robot trajectory.
The original capsule model (shown in red) was generated using cornea, iris, and PC data from the transpupillary OCT.
The scanning trajectory was then generated on the capsule model subject to constraints of joint angles and tissue positions.
Using intraocular measurements of the capsule in the equatorial region, the capsule model was updated (shown in green), which accounts for modeling errors due to insufficient visualization of the equatorial region from the transpupillary OCT.
Based on intraocular A-scan measurements, the modeling error without intraocular measurements (shown in red) is calculated to be approximately 596 \textmu m.
Note that the IOP control was activated throughout the procedure to maintain the natural shape of the eye.

\subsection{Capsule Cleaning with Intraocular OCT}

\begin{figure}[t!]
    \centering
    \includegraphics[width=0.7\linewidth]{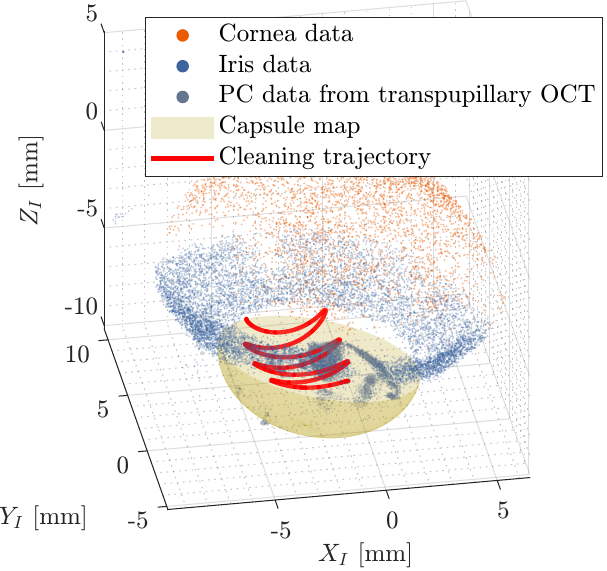}
    \caption{Shown in red is an example of the cleaning trajectory that covers both the pupillary and equatorial region.}
    \label{fig:cleaning_trajectory}
\end{figure}

\begin{figure}[t!]
    \centering
    \includegraphics[width=0.85\linewidth]{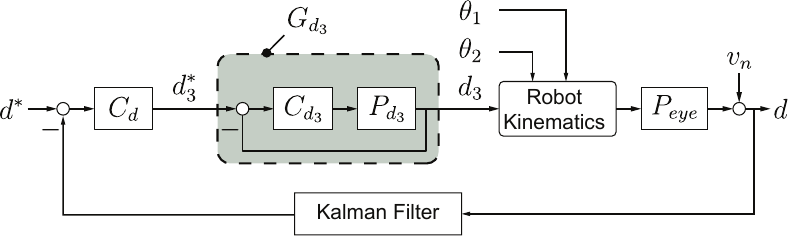}
    \caption{Shown is the inner- and outer-loop control strategy for A-scan feedback compensation on the feedforward trajectory. $G_{d_3}$ is the inner-loop PID control for joint 3 and $C_d$ is the outer-loop PI control to track the safe distance $d^*$.}
    \label{fig:ascan_fb_diagram}
\end{figure}

In \cite{lee2023accurate}, a sinusoidal scan trajectory was selected due to its short elapsed time and adequate capsule coverage.
Fig. \ref{fig:cleaning_trajectory} shows an example of the sinusoidal trajectory generated within the boundary of the constructed capsule map. 
To account for tissue movement during polishing, an additional outer loop controller is added to track distance in the tool insertion actuator to satisfy the distance constraint while performing the surgery.
Although it is desirable to detect three-dimensional distances, we develop the feedback controller on a single joint ($d_3$ in Fig. \ref{fig:robot_coordinate_system}) because the intraocular probe provides only the axial distance information.
As shown in Fig. \ref{fig:ascan_fb_diagram}, the controller contains an inner loop and an outer loop.
The inner loop controller ($G_{d_3}$) is the closed-loop motor PID controller which runs at 1 kHz and has a bandwidth of 18 Hz.
The outer loop uses the A-scan distance measurements $d$ as feedback.
A single A-scan is calculated by averaging over 100 intensity graphs to reduce background noise, reaching a data acquisition rate at approximately 20 Hz. 
From each A-scan, $d$ is calculated by the edge detection algorithm.
A PI controller was designed to perform distance tracking initialized with the Ziegler-Nichols method.
By accounting for the data process time and the robot dynamics, the outer loop distance tracking controller runs at a slower rate of 10 Hz.

\begin{figure}[t!]
    \centering
    \includegraphics[width=0.75\linewidth]{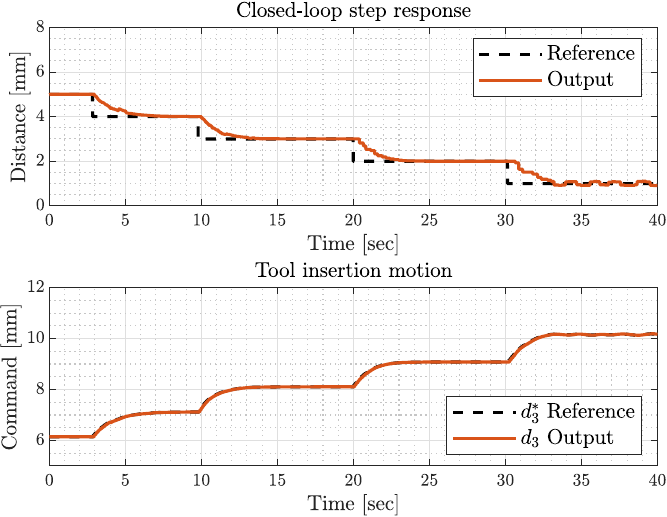}
    \caption{Shown is the closed-loop step response. Top: Distance step tracking response with different distance set points. Bottom: Tool insertion motion ($d_3$) adjustments based on the distance feedback.}
    \label{fig:ascan_cl_step_resp}
\end{figure}

\begin{figure}[t!]
    \centering
    \includegraphics[width=0.75\linewidth]{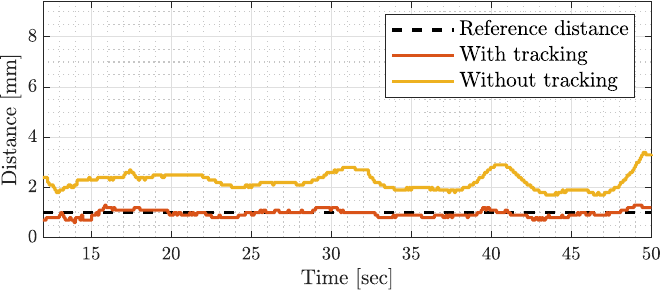}
    \caption{Shown is the trajectory tracking performance with a distance set point at 1 mm during capsule cleaning.}
    \label{fig:ascan_cleaning_tracking}
\end{figure}

The closed-loop step response is shown in Fig. \ref{fig:ascan_cl_step_resp} with seven incremental step commands from 5 to 1 mm, showing its ability to track different reference distances.
The rise time is approximately 1 s with zero steady-state error.
The distance feedback performance is also evaluated by changing $\theta_1$ and $\theta_2$ with a section of the cleaning trajectory with $-60^\circ < \theta_1 < -40^\circ$ and $0^\circ < \theta_2 < 20^\circ$.
As shown in Fig. \ref{fig:ascan_cleaning_tracking}, the trajectory tracking with distance control successfully maintained the distance at a set point of 1 mm with a mean of 47 \textmu m, while the tool to tissue distance without control has a mean error of 763 \textmu m. 
However, if there is modeling error, the instrument may accidentally be in contact with the tissue.

\subsection{Post-Operative Analysis}

\begin{figure}[t!]
    \centering
    \includegraphics[width=\linewidth]{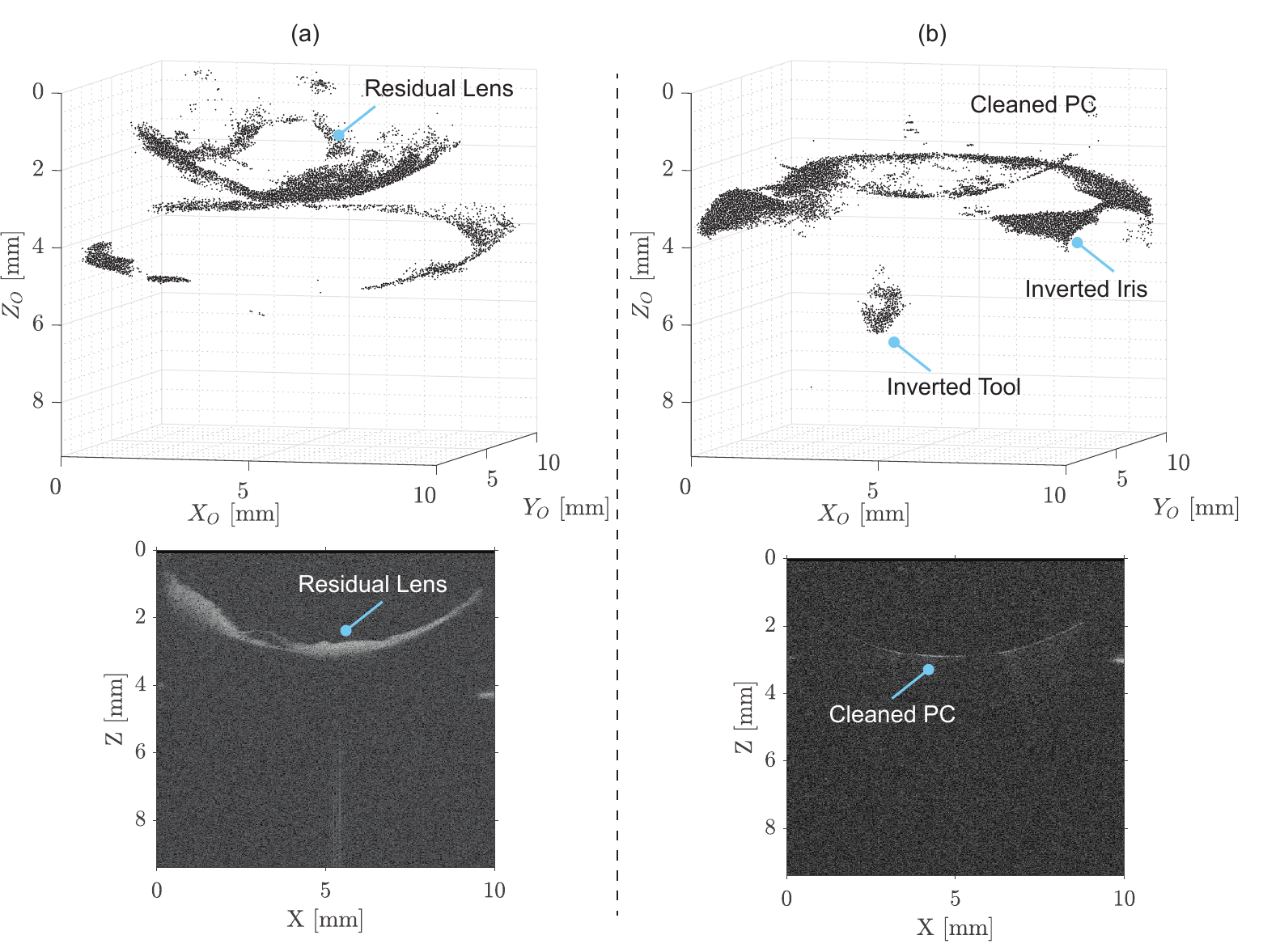}
    \caption{Volume scans and B-scans to show preoperative (left) and postoperative (right) comparison. The PC was cleaned and intact after capsule cleaning.}
    \label{fig:polishing_compare}
\end{figure}

\begin{figure}[t!]
    \centering
    \includegraphics[width=\linewidth]{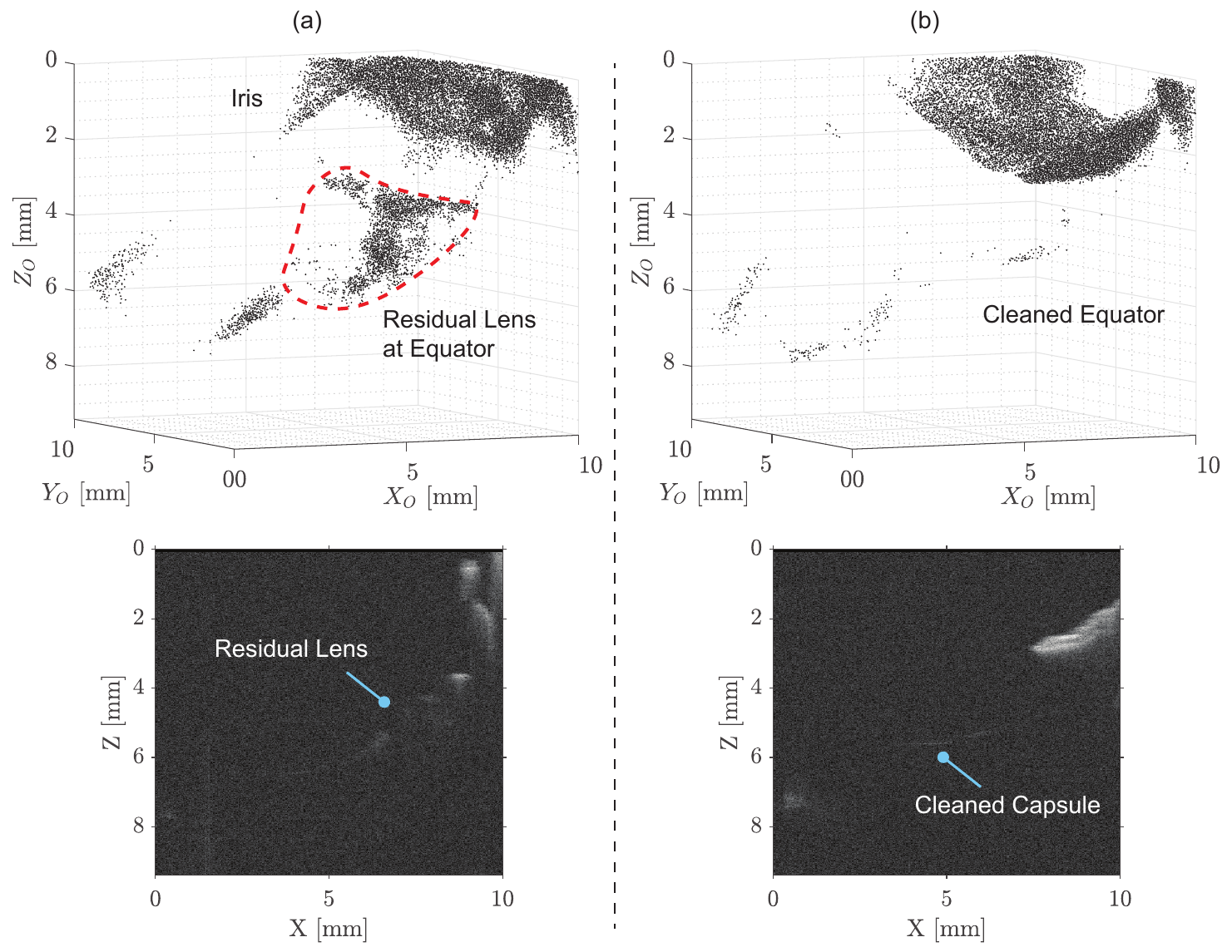}
    \caption{Shown is the preoperative and postoperative comparison when tilting the eye 45$^\circ$ to assess capsule cleaning in the equatorial region.}
    \label{fig:polishing_compare_tilt}
\end{figure}

Capsule cleaning was performed on three \textit{ex-vivo} pig eyes.
At the end of the procedure, the integrity of the tissues (cornea, iris, and capsule) was assessed by a trained surgeon using a digital microscope.
Fig. \ref{fig:polishing_compare} illustrates the effectiveness in cleaning the pupillary capsule, where no residual lens materials were observed.
To assess the cleanliness of the equatorial region, the whole eye was tilted 45$^\circ$ and ophthalmic viscosurgical devices were injected into the anterior chamber to improve visualization under OCT.
The residual lens on the peripheral region was removed as shown in Fig. \ref{fig:polishing_compare_tilt}.
The average cleaning procedure time was 55 s, which was longer than the pupillary polishing in our previous work in \cite{lee2023accurate} due to the larger capsule coverage, but still presents advantages over safe distance regulation and equatorial cleaning.
No tissue damage was observed in the three trials.

% =============================================

\section{Conclusion} \label{sec:conclusion}

\noindent \hl{We presented an innovative capsule mapping approach that uses transpupillary OCT for eye registration and intraocular OCT to ensure the modeling accuracy of the lens capsule.
Capsule mapping on the eye phantom and capsule cleaning on \textit{ex-vivo} pig eyes are summarized below:}
\begin{itemize}
    \item \hl{Capsule localization: Our method demonstrated improved equatorial capsule modeling with intraocular scans over completely relying on transpupillary OCT and achieved an accuracy of 53 \textmu m}. 
    \item \hl{Accurate registration: We have intra-operatively established the eye position using transpupillary OCT without the prior knowledge of eye position for robotic intraocular scanning.}
    \item \hl{In-situ calibration: With robot precision, correct optical refractive indices can be online identified (2.3\% error compared to the ground truth), and a reduction of 88\% in capsule modeling error.}
    \item \hl{Capsule cleaning: With the constructed capsule map and real-time A-scan distance feedback, the cleaning procedure ensures complete residual lens removal without tissue damage.}
\end{itemize}

\hl{It is expected that the developed technology will facilitate the polishing procedure by enhancing equatorial visualization with accuracy guaranteed.
Although the effectiveness of the developed method was demonstrated, future work will focus primarily on evaluating the method on larger biological samples with more variations.
To achieve this, the following steps are outlined to address the foreseen challenges.}

\hl{First, the scanning trajectory was pre-determined and generated with prior assumptions of the capsule anatomy, but an optimal trajectory subject to tooltip travel time, robot joint limits, dynamics, and tissue positions remains to be explored.
While we chose a relatively large OCT volume scan width and the present acquisition delay, microscopic images can be used to locate the OCT in the pupillary region and perform a smaller OCT scan to accelerate the localization process.
Finally, the improved capsule modeling and visualization approach presented in this paper may also be applied to peripheral vitreous shaving or other surgical procedures that require visualization of blocked tissues while allowing tissue registration from a visible area.}

\bibliographystyle{IEEEtran}
\bibliography{reference}

\end{document}